%% file: main.tex
\documentclass{article} 
\usepackage{iclr2024_conference,times}
\input{math_commands.tex}

\usepackage[hidelinks]{hyperref}
\usepackage{url}

\usepackage{array}
\usepackage{amsfonts}       
\usepackage{amsmath}
\usepackage{amssymb}
\usepackage[utf8]{inputenc} 
\usepackage[T1]{fontenc}   
\usepackage{pifont}
\newcommand{\xmark}{--} 
 
\usepackage[algoruled,algo2e]{algorithm2e}
\usepackage{mathtools}
\usepackage{wasysym}

\usepackage[compact]{titlesec}

\titlespacing{\subsection}{0pt}{0.2\parskip}{-0.05\parskip}
\titlespacing{\subsubsection}{0pt}{0.2\parskip}{-0.1\parskip}

\usepackage{bookmark}

\usepackage{wrapfig,lipsum,booktabs}
\usepackage[bottom]{footmisc}
\usepackage{makecell}
\usepackage{tabularx}
\usepackage{booktabs}
\usepackage{multirow}

\usepackage{nicefrac}       %
\usepackage{xfrac}
\usepackage{microtype}      %

\usepackage{xcolor}         %
\usepackage{graphicx}
\graphicspath{{figures/}}
\usepackage{subcaption}
\usepackage{caption}
\usepackage{svg}
\usepackage{capt-of}
\usepackage{varwidth}

\DeclareRobustCommand{\acro}[1]{\textsc{\lowercase{#1}}}
\usepackage{bm}
\usepackage[normalem]{ulem}
\useunder{\uline}{\ul}{}

\DeclareSymbolFont{symbolsC}{U}{txsyc}{m}{n}
\SetSymbolFont{symbolsC}{bold}{U}{txsyc}{bx}{n}
\DeclareFontSubstitution{U}{txsyc}{m}{n}
\DeclareMathSymbol{\multimapboth}{\mathrel}{symbolsC}{"13}
\usepackage[shortlabels]{enumitem}

\newcommand*\samethanks[1][\value{footnote}]{\footnotemark[#1]}

\title{Predictive, scalable and interpretable knowledge tracing on structured domains}

\author{Hanqi Zhou$^{1,2,4}$, Robert Bamler$^{1,3}$, Charley M. Wu$^{1,2,3}$\thanks{Equal contribution. Code at \href{https://github.com/mlcolab/psi-kt}{github.com/mlcolab/psi-kt}}, \& Álvaro Tejero-Cantero$^{1,2}$\samethanks{} \\
$^1$University of Tübingen, 
$^2$Cluster of Excellence Machine Learning, 
$^3$Tübingen AI Center, 
$^4$IMPRS-IS \\ 
\texttt{\{hanqi.zhou,robert.bamler,charley.wu,alvaro.tejero\}@uni-tuebingen.de} \\ %
}

\iclrfinalcopy %

\begin{document}
\maketitle

\setlength{\abovedisplayskip}{2pt}
\setlength{\belowdisplayskip}{2pt}
\setlength{\abovedisplayshortskip}{2pt}
\setlength{\belowdisplayshortskip}{2pt}

\begin{abstract}
Intelligent tutoring systems optimize the selection and timing of learning materials to enhance understanding and long-term retention. 
This requires estimates of both the learner's progress (``knowledge tracing''; KT), and the prerequisite structure of the learning domain (``knowledge mapping''). 
While recent deep learning models achieve high KT accuracy, they do so at the expense of the interpretability of psychologically-inspired models.
In this work, we present a solution to this trade-off. 
PSI-KT is a hierarchical generative approach that explicitly models how both individual cognitive traits and the prerequisite structure of knowledge influence learning dynamics, thus achieving interpretability by design. %
Moreover, by using scalable Bayesian inference, PSI-KT targets the real-world need for efficient personalization even with a growing body of learners and learning histories. %
Evaluated on three datasets from online learning platforms, PSI-KT achieves superior multi-step \textbf{p}redictive accuracy and \textbf{s}calable inference in continual-learning settings, all while providing \textbf{i}nterpretable representations of learner-specific traits and the prerequisite structure of knowledge that causally supports learning. 
In sum, predictive, scalable and interpretable knowledge tracing with solid knowledge mapping lays a key foundation for effective personalized learning to make education accessible to a broad, global audience.%
\end{abstract}

\input{main_text/1_introduction} %
\input{main_text/2_background} %
\input{main_text/3_method} %
\input{main_text/4_datasets} %
\input{main_text/5_1_exp_pred} %
\input{main_text/5_2_exp_learner}
\input{main_text/5_3_exp_graph} %

\input{main_text/6_discussion}

\subsubsection*{Acknowledgments}
The authors thank Nathanael Bosch and Tim Z. Xiao for their helpful discussion, and Seth Axen for code review. 
The authors thank the International Max Planck Research School for Intelligent Systems (IMPRS-IS) for supporting Hanqi Zhou. 
This research was supported as part of the LEAD Graduate School \& Research Network, which is funded by the Ministry of Science, Research and the Arts of the state of BadenWürttemberg within the framework of the sustainability funding for the projects of the Excellence Initiative II.
Funded by the Deutsche Forschungsgemeinschaft (DFG, German Research Foundation) under Germany’s Excellence Strategy – EXC number 2064/1 – Project number 390727645.
CMW is supported by the German Federal Ministry of Education and Research (BMBF): Tübingen AI Center, FKZ: 01IS18039A.

\subsubsection*{Ethics statement}
We evaluated our \acro{PSI-KT} model on three public datasets from human learners, which all anonymize the data to protect the identities of individual learners. Although \acro{PSI-KT} aims to improve personalized learning experiences, it infers cognitive traits from behavioral data instead of using learners' demographic characteristics (e.g., age, gender, and the name of schools provided in the Assistment 17 dataset), to avoid reinforcing existing disparities.

Evaluations of structured knowledge tracing in our paper are limited by dataset availability to pre-college mathematics. To ensure a broader and more ecologically valid assessment, it is essential to explore diverse datasets across various domains (e.g., biology, chemistry, linguistics) and educational stages (from primary to college level). This will allow for a more comprehensive understanding of the role of structure in learning.  

\bibliography{iclr2024_conference}
\bibliographystyle{iclr2024_conference}

\newpage
\appendix
\input{appendix/7_1_elbo} %
\input{appendix/7_2_baseline} %
\input{appendix/7_3_groupkt_model} %
\input{appendix/7_4_exp_pred}

\input{appendix/7_5_exp_learner}
\input{appendix/7_6_exp_graph}

\input{appendix/7_7_ablation_study}

\end{document}

%% file: math_commands.tex
\usepackage{amsmath,amsfonts,bm}

\newcommand{\figleft}{{\em (Left)}}

\newcommand{\figright}{{\em (Right)}}
\newcommand{\figtop}{{\em (Top)}}
\newcommand{\figbottom}{{\em (Bottom)}}
\newcommand{\captiona}{{\em (a)}}
\newcommand{\captionb}{{\em (b)}}
\newcommand{\captionc}{{\em (c)}}
\newcommand{\captiond}{{\em (d)}}
\newcommand{\captione}{{\em (e)}}
\newcommand{\captionf}{{\em (f)}}
\newcommand{\captiong}{{\em (g)}}

\def\eqref#1{equation~\ref{#1}}

\def\1{\bm{1}}

\DeclareMathAlphabet{\mathsfit}{\encodingdefault}{\sfdefault}{m}{sl}
\SetMathAlphabet{\mathsfit}{bold}{\encodingdefault}{\sfdefault}{bx}{n}

\newcommand{\E}{\mathbb{E}}

%% file: main_text/1_introduction.tex
\section{Introduction}
\label{sec:introduction}
The rise of online education platforms has created new opportunities for personalization in learning, motivating a renewed interest in how humans learn structured knowledge domains.
Foundational theories in psychology \citep{ebbinghaus1913memory} have informed \emph{spaced repetition} schedules \citep{kt-hlr}, which exploit the finding that an optimal spacing of learning sessions enhances memory retention. %
Yet beyond the timing of rehearsals, the sequential order of learning materials is also crucial, as evidenced by curriculum effects in learning \citep{dewey1910child, dekker2022curriculum}, where exposure to simpler, prerequisite concepts can facilitate the apprehension of higher-level ideas. %
Cognitive science and pedagogical theories have long emphasized the relational structure of knowledge in human learning \citep{rumelhart2017schemata, piaget1970science}, with recent research showing that mastering prerequisites enhances concept learning \citep{lynn2020humans, karuza2016local, brandle2022exploration}. %
Yet, we still lack a predictive, scalable, and interpretable model of the structural-temporal dynamics of learning that could be used to develop future intelligent tutoring systems. 

Here, we present \acro{PSI-KT}, a novel approach for inferring interpretable learner-specific cognitive traits and a shared knowledge graph of prerequisite concepts. We demonstrate our approach on three real-world educational datasets covering structured domains, where our model outperforms existing baselines in terms of \emph{predictive} accuracy (both within- and between-learner generalization), \emph{scalability} in a continual learning setting, and \emph{interpretability} of learner traits and prerequisite graphs. 
The paper is organized as follows: We first introduce the knowledge tracing problem and summarize related work (Sec.~\ref{sec:background}). We then provide a formal description of \acro{PSI-KT} and describe the inference method (Sec.~\ref{sec:model}). Experimental evaluations are organized into demonstrations of prediction performance, scalability, and interpretability (Sec.~\ref{sec:results}).
Altogether, \acro{PSI-KT} bridges machine learning and cognitive science, leveraging our understanding of human learning to build the foundations for automated tutoring systems with broad educational applications.

%% file: main_text/2_background.tex
\section{Background} \label{sec:background}
In this section, we begin by defining the knowledge tracing problem and then review related work. 
\subsection{Knowledge tracing for intelligent tutoring systems} \label{sec:background-kt}
For almost 100 years \citep{pressey1926simple}, researchers have developed intelligent tutoring systems (\acro{ITS}) to support human learning through adaptive teaching materials and feedback. 
More recently, \emph{knowledge tracing} \citep[KT;][]{corbett1994knowledge} emerged as a method for tracking learning progress by predicting a learner's performance on different \emph{knowledge components} (KCs), e.g., the `Pythagorean theorem', based on past learning interactions. 
Here, we focus on the KT problem, with the goal of supporting the selection of teaching materials in future ITS applications.

In this setting, a learner~$\ell$ receives exercises or flashcards for KCs~$x^\ell_n \in \{0, 1, \ldots, K\}$ at irregularly spaced times~$t_n^\ell$, whereupon the performance is recorded, often as correct/incorrect,~$y^\ell_n \in \{0,1\}$. We can formalize KT as a supervised learning problem on time-series data, where the goal of the KT model is to predict future performance (e.g.,~$\hat{y}_{N+1}$) given all or part of the interaction history~$\mathcal{H}_{1:N}^\ell{:=}\{x_n^\ell, t_n^\ell, y_n^\ell \}_{n=1}^N$ available up to time~$t_N^\ell$.  As part of the process, a KT model may infer specific representations of learners or of the learning domain to help prediction.
If these representations are interpretable, they can be valuable for downstream learning personalization.

\subsection{Related work} \label{sec:related-work}
We broadly categorize related KT approaches into psychological and deep learning methods.

\textbf{Psychological methods. }
Focusing on interpretability, psychological methods use domain knowledge to describe the temporal decay of memory \citep[e.g., forgetting curves;][]{ebbinghaus1913memory}, sometimes also modeling learner-specific characteristics.
\noindent\emph{Factor-based regression} models use hand-crafted features based on learner interactions and KC properties \citep[e.g., repetition counts and KC easiness;][]{kt-pfa}. While they model KC-dependent memory dynamics \citep{kt-pavlik2021logistic, kt-gervet2020deep, kt-lindsey2014improving, kt-irt, kt-mirt}, they ignore the relational structure between KCs. 
Half-life Regression \citep[\acro{HLR};][]{kt-hlr} from Duolingo uses both correct and incorrect counts, 
while the Predictive Performance Equation \citep[\acro{PPE};][]{kt-ppe} models the elapsed time of every past interaction with a power function to account for spacing effects.
By using shallow regression models with predefined features, these models achieve interpretability, but sacrifice prediction accuracy. 
\noindent\emph{Latent variable models} use a probabilistic two-state Hidden Markov Model \citep{kt-kaser2017dynamic, kt-sao2013incorporating, kt-baker2008more, kt-yudelson2013individualized}, representing either mastery or non-mastery of a given KC. These models are limited to binary states by design, do not account for learner dynamics, and for some, their numerous parameters hinder scalability.  
Another probabilistic model, \acro{HKT} \citep{kt-hkt} accounts for structure and dynamics by modeling knowledge evolution as a multivariate Hawkes process. Close in spirit to our \acro{PSI-KT}, this approach tracks KC structure but lacks any learner-specific representations.

\textbf{Deep learning methods. }
Deep learning methods use flexible models with many parameters to achieve high prediction accuracy. However, this flexibility also makes it difficult to interpret their learned internal representations.
The first deep learning methods explicitly modeled sequential interactions with \noindent\emph{recurrent neural networks} to overcome the dependence on fixed summary statistics in simpler regression models, with
Deep Knowledge Tracing \citep[\acro{DKT};][]{kt-dkt} pioneering the use of Long Short-Term Memory (LSTM) networks \citep{lstm}.
A similar architecture, \acro{DKTF} \citep{kt-dktf} incorporated additional input features, whereas \citet{kt-lpkt} proposed an intricate modular architecture aimed at recovering interpretable learner representations, but neglecting KC relations.  
\noindent\emph{Structure-aware models} leverage KC dependencies, accounting for the fact that human knowledge acquisition is structured by dependency relationships \citep[i.e., concept maps;][]{hill2005concept, koponen2018concept, lynn2020humans}.
\citet{kt-skt} empirically estimate KC dependencies from the frequencies of successful transitions. \acro{AKT} \citep{kt-akt} relies on the attention mechanism \citep{transformer} to implicitly capture structure \citep{kt-sakt, kt-saint, kt-saint+, kt-simple}, whereas \acro{GKT} \citep{kt-gkt} models it explicitly based on graph neural networks \citep{gcn}. %
Recent work towards interpretable deep learning KT uses engineered features such as learner mastery and exercise difficulty \citep{kt-ikt}, or infers them with neural networks (\citealp[\acro{QIKT};][]{kt-qikt}, \citealp[\acro{IEKT};][]{kt-iekt}). While diverse approaches to interpretability exist (\citealp[see][for review]{kt-qikt}), a comprehensive evaluation framework is still lacking.

Here, we present our predictive, scalable and interpretable KT model (\acro{PSI-KT}) as a psychologically-informed probabilistic deep learning approach, together with a comprehensive evaluation framework for interpretability. 

%% file: main_text/3_method.tex
\section{Joint dynamical and structural model of learning} \label{sec:model}

\begin{figure}[t]
\vspace{-10pt}
    \centering
    {\includegraphics[height=1.4in]{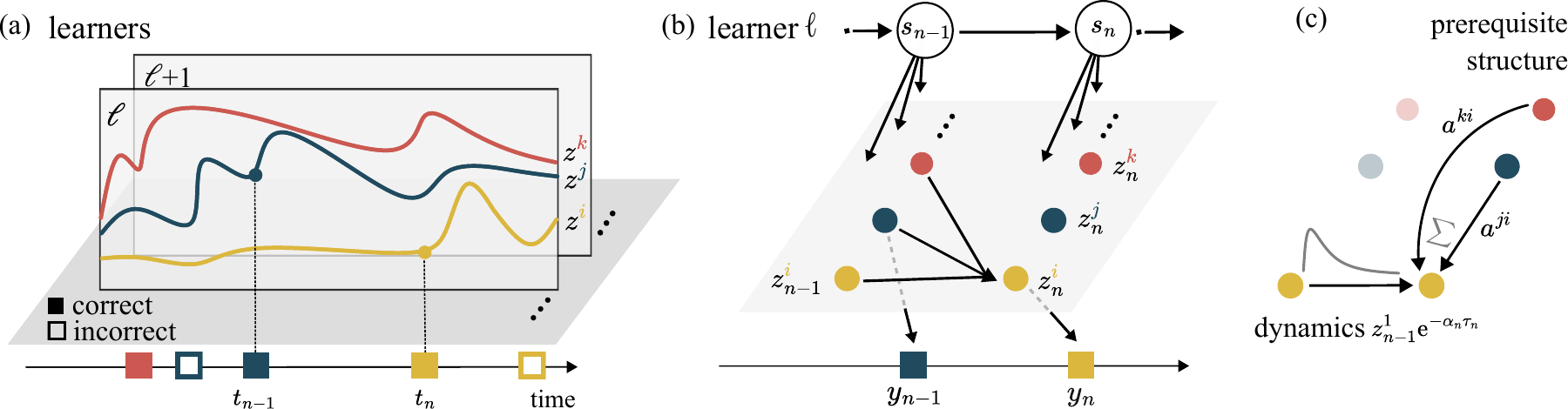}} 
    \caption{
    \acro{PSI-KT} is a hierarchical probabilistic state-space model of learning.
    \captiona~Latent knowledge states for different KCs (colored curves) are inferred from observations. 
    \captionb~Full hierarchical model for a single learner: cognitive traits $s_n$ control the coupled dynamics of states $z^k_n$, which give rise to observations $y_n$.
    \captionc~The dynamics combine memory decay (Eq.~\ref{eq:ou-process-marginal}) and structural influences (Eq.~\ref{eq:ou-process-mean}).
    }
    \label{fig:model}
\vspace{-10pt}
\end{figure}

In this section, we describe \acro{PSI-KT}, our probabilistic hierarchical state-space model of human learning (Fig.~\ref{fig:model}). 
Briefly, observations of learner performance~$y$ (Fig.~\ref{fig:model}a, filled/unfilled boxes) provide indirect and noisy evidence about latent knowledge states~$z$ (colored curves, with matching dots in Fig.~\ref{fig:model}b). 
These latent states evolve stochastically, in line with the psychophysics of memory (temporal decay in Fig.~\ref{fig:model}c), while simultaneously being subject to structural influences from performance on prerequisite KCs (structure in Fig.~\ref{fig:model}c). 
We introduce a second latent level of learner-specific traits~$s$ (Fig.~\ref{fig:model}b, top), which govern the knowledge dynamics in an interpretable way. 

Below, we describe the method in more detail.
We start with the generative model (Sec.~\ref{sec:generative-model}).
Next, we discuss the joint approximate Bayesian inference of latent variables and estimation of generative parameters (Sec.~\ref{sec:inference}).
Finally, we show how to derive multi-step performance predictions (see Sec.~\ref{sec:predictions} and Fig.~\ref{appfig:model_detail} in Appendix~\ref{appsec:groupkt-architecture} for a graphical overview of inference and prediction).

\subsection{Probabilistic state-space generative model}  \label{sec:generative-model}
We conceptualize observations of learner performance as noisy measurements of an underlying time-dependent knowledge state, specific to each learner and KC. The evolution of knowledge states reflects the process of learning and forgetting, governed by learner-specific traits. Additionally, knowledge of different KCs informs one another according to learned prerequisite relationships.
We translate these modeling assumptions into a generative model consisting of three main components:%
(i)~the learner knowledge state across KCs,~$\bm{z}^\ell_n = [z_n^{\ell, 1} \ldots z_n^{\ell, K}]^{\intercal} \in \mathbb{R}^K$ (colored curves in Fig.~\ref{fig:model}a),
(ii)~learner-specific cognitive traits~$s^\ell_n \in \mathbb{R}^4$ (top row in Fig.~\ref{fig:model}b), and
(iii)~a shared static graph~$\mathcal{A}$ of KCs whose edges~$a^{ik}$ quantify the probability for a KC~$i$ to be a prerequisite for KC~$k$ (Fig.~\ref{fig:model}c).

\paragraph{State-space model. } State-space models (SSMs) are a framework for partially observable dynamical processes.
They represent the inherent noise of measurements~$y$ by an emission distribution~$p (y_n \,|\, z_n)$, separate from the stochasticity of state dynamics, modeled as a first-order Markov process with transition probabilities~$p (z_n \,|\, z_{n-1})$. The state dynamics are initiated by sampling from an initial prior~$p (z_1)$ to iteratively feed the transition kernel, and predictions can be drawn at any time from the emission distribution. 
To represent the influence of individual cognitive traits over the knowledge dynamics, we additionally condition the $z$-transitions on the traits~$s$ (which also can be observed only indirectly). 
The three-level SSM hierarchy of \acro{PSI-KT} consists of:
\vspace{0.2cm}
\begin{alignat}{2}
  &\textrm{Level 2 (latent cognitive traits):}\quad&s_n^\ell &\sim p_\theta (s^\ell_n \,|\, s^\ell_{n - 1}):=\mathcal{N} (s^\ell_n \,|\,  Hs^\ell_{n - 1}, R) \label{eq:ssm-traits} \\
  &\textrm{Level 1 (latent knowledge states):} \quad& \bm{z}^\ell_n &\sim  p_\theta (\bm{z}^\ell_n \,|\, \bm{z}^\ell_{n - 1}, s^\ell_n):=\textstyle{\prod_k}\ \mathcal{N} (z^{\ell, k}_n  | m^{\ell, k}_n, w^\ell_n) \label{eq:ssm-states} \\
  &\textrm{Level 0 (observed learner performance):} \quad& \hat{y}^\ell_n &\sim p  (y^\ell_n \,|\, z^{\ell, k}_n) := \mathrm{Bern} (\operatorname{sigmoid} (z^{\ell, k}_n)) . \label{eq:ssm-observations}
\end{alignat}
The choice of Gaussian initial priors (discussed below) and Gaussian transitions ensures tractability, while the Bernoulli emissions model the observed binary outcomes. We now unpack this model and all its parameters in detail, starting with the knowledge dynamics.

\paragraph{Knowledge states $\bm{z}$.} Recent KT methods \cite[e.g.,][]{kt-dktf} use an exponential forgetting function based on psychological theories \citep{ebbinghaus1913memory}. Here, we augment this approach by adding stable long-term memory~{\citep{averell2011form}}, and model the knowledge dynamics~$z^{\ell, k}$ of an isolated KC~$k$ as a mean-reverting stochastic (Ornstein-Uhlenbeck; OU) process:
\begin{align} \mathrm{d} z^{\ell, k}/ \mathrm{d} t = \alpha^\ell  (\mu^\ell - z^{\ell, k})
  + \sigma^\ell \eta (t). \label{eq:langevin} \end{align}
Accordingly, the state of knowledge~$z^\ell$ gradually reverts to a long-term mean~$\mu^\ell$ with rate~$\alpha^\ell$, subject to white noise fluctuations~$\eta(t)$ scaled by volatility~$\sigma^\ell$. To account for the influence of other KCs, we adjust the mean~$\mu^\ell_n$ using prerequisite weights~$a^{ik}$ (defined in Eq.~\ref{eq:adjacency-matrix} below), modulated by the learner's transfer ability~$\gamma^\ell_n$:
\begin{align}  
\tilde{\mu}^{\ell, k}_n &:= \mu^\ell_n + (\gamma^\ell_n/K) 
   \textstyle \sum_{i \neq k} a^{ik}\,z_n^{\ell, i}. 
  \label{eq:ou-process-mean}
\end{align}
We obtain the mean~$m_n^{\ell,k}$ and variance~$w_n^\ell$
of the transition kernel in Eq.~\ref{eq:ssm-states} by marginalizing the OU process over one time step~$\tau_n^\ell := t_n^\ell - t_{n - 1}^\ell$, which can be done analytically\footnote{\cite{sarkka2019applied} --- the variance is~$w^\ell_n= (\sigma^\ell_n)^2 (1-{\rm e}^{-2\alpha^\ell_n\tau^\ell_n})/(2\alpha^\ell_n)$.} ,
\begin{align}
  m^{\ell, k}_n &= r_n^\ell\, z^{\ell, k}_{n - 1}\ +\ (1 - r_n^\ell)\,\tilde{\mu}^{\ell, k}_n, \quad \text{with retention ratio} \  
    r_n^\ell := \mathrm{e}^{- \alpha^\ell_n \tau^\ell_n} \in (0, 1).  \label{eq:ou-process-marginal}
\end{align}
As the time since the last interaction~$\tau_n^\ell$ grows, the retention ratio~$r^\ell_n$ decreases exponentially with rate~$\alpha^\ell_n$, and the knowledge state reverts to the long-term mean~$\tilde\mu_n^{\ell,k}$, which partly depends on the learner's mastery of prerequisite KCs (Eq.~\ref{eq:ou-process-mean}).
This balances short-term and long-term learning, reflecting empirical findings from memory research \citep{averell2011form}.
The structural influences are accounted for in the dynamics of~$z_n^{\ell,k}$, thus justifying the conditional independence assumed in Eq.~\ref{eq:ssm-states}. A Gaussian initial prior~$p_{\theta} (z_1^{\ell, k}) =\mathcal{N} (z^{\ell, k}_1|\bar{z}, w_1)$, where~$\bar{z}, w_1 \in \mathbb{R}$ are part of the generative parameters~$\theta$, completes our dynamical model of knowledge states.

\paragraph{Learner-specific cognitive traits $s$.} The dynamics of knowledge states (Eqs.~\ref{eq:langevin}-~\ref{eq:ou-process-marginal}) are parameterized by learner-specific
cognitive traits~$(\alpha^\ell_n, \mu^\ell_n, \sigma^\ell_n, \gamma^\ell_n)$, which we collectively denote~$s^\ell_n$. Specifically, $\alpha^\ell$ represents the forgetting rate \citep{ebbinghaus1913memory,averell2011form}, $\mu^\ell$ (via $\tilde{\mu}^{\ell,n}_k$) captures long-term memory consolidation \citep{meeter2004consolidation} for practiced KCs and expected performance for novel KCs, $\sigma^\ell$ quantifies knowledge volatility, and $\gamma^\ell$ measures transfer ability \citep{bassett2017network} from knowledge of prerequisite KCs.
These traits can develop during learning according to Eq.~\ref{eq:ssm-traits}, starting from a Gaussian prior~$p_{\theta} (s^\ell_1) =\mathcal{N} (s^\ell_1|\bar{s}, R_1)$ where~$\bar{s} \in \mathbb{R}^4$ and the diagonal matrices~$H, R_1, R \in \mathbb{R}^{4 \times 4}$ are also part of the global parameters~$\theta$. 

\paragraph{Shared prerequisite graph $\mathcal{A}$.} 
In our model, prerequisite relations influence knowledge dynamics via the coupling introduced in Eq.~\ref{eq:ou-process-mean}. We now discuss an appropriate parameterization for the weight matrix of the prerequisite graph,~$ \mathcal{A}:=\{a^{ik}\}_{i,k \in 1:K}$. We assume that prerequisites are time- and learner-independent so that, in the spirit of collaborative filtering \citep{breese2013empirical}, we can pool evidence from all learners to estimate them. To prevent a quadratic scaling in the number of KCs, we do not directly model edge weights but derive them from KC embedding vectors~$u^k$ in lower dimension~$u^k \in \mathbb{R}^D$ with~$D \ll K$, collected in embedding matrix~$U_{K\times D}$.
A basic integrity constraint for a connected pair is that dependence of KC~$i$ on KC~$k$ should trade off against that of~$k$ on~$i$, i.e., no mutual prerequisites: $a^{ik} + a^{ki}=1$. With this in mind, we exploit the factorization of~$a^{ik}$ introduced by \cite{lippe2021efficient} in terms of a separate probability of edge existence~$p(i \multimapboth\! k)$ and definite directionality~$p(i\!\rightarrow\! k \,|\, i \multimapboth\! k)$:
\begin{align}\label{eq:adjacency-matrix}
        a^{ik} &:= p(i\!\rightarrow\! k \,|\, i \multimapboth\! k)\, p(i\multimapboth\! k) \nonumber\\
               & = \operatorname{sigmoid}((u^i)^\intercal u^k)\, \operatorname{sigmoid}((u^i)^\intercal (M-M^\intercal)\, u^k),
\end{align}
where the skew-symmetric combination~$M-M^\intercal$ of a learnable matrix~$M$ prevents mutual prerequisites.
Having presented the generative model, we now turn to inference and prediction.

\subsection{Approximate Bayesian Inference and Amortization with a Neural Network} \label{sec:inference}
We now describe how we learn the generative model parameters~$\theta$ and how we infer the latent states~$s, z$ introduced in Section~\ref{sec:generative-model} using a neural network (``inference network'').
Since learner-specific latent states~$s$ and~$z$ are deducible solely from limited individual data, we expect non-negligible uncertainty. This motivates our probabilistic treatment of these states using approximate Bayesian inference.
By contrast, the model parameters~$\theta$ (KC parameters~$U, M$ in Eq.~\ref{eq:adjacency-matrix}, transition parameters~$\bar s, H, R_1, R$ in Eq.~\ref{eq:ssm-traits}, and~$\bar z, w_1$ in Eq.~\ref{eq:ssm-states}) can be estimated from all learners, and we thus treat them as point-estimated parameters as described below (detailed derivation in Appendix~\ref{appsec:elbo}.) Here, without loss of generality, we show the inference for a single learner.

\subsubsection{Inference on a fixed learning history} \label{sec:inference-entire-history}
Here, we assume the full interaction history~$\mathcal{H}_{1:N}^\ell$ is available for inferring the posterior over latents~$p_\theta(\bm{z}_{1:N}^\ell, s_{1:N}^\ell \,|\, y_{1:N}^\ell)$.
We approach the problem using variational inference (VI). In VI, we select a distribution family~$q_\phi$ with free parameters~$\phi$ to approximate the posterior~$p_\theta$ by minimizing their Kullback-Leibler divergence. This can only be done indirectly, by maximizing a lower bound to the marginal probability of the data, the \emph{evidence lower bound} ($\rm ELBO$). Here, we adopt the mean-field approximation~$q_\phi(\bm{z}_{1:N}^\ell, s_{1:N}^\ell \,|\, y_{1:N}^\ell)=q_\phi(\bm{z}_{1:N}^\ell)\,q_\phi(s_{1:N}^\ell)$ and jointly optimize the generative~$\theta$ and variational~$\phi$ parameters using variational \emph{expectation maximization} \citep[EM;][]{dempster1977maximum, beal2003variational, attias1999variational}. Motivated by real-world scalability, we introduce an inference network (see Appendix \ref{appsec:baseline-model} for the architecture) to amortize the learning of variational parameters~$\phi$ across learners, and we employ the \emph{reparametrization trick}~\citep{kingma2014autoencoding} to optimize the single-learner~$\rm ELBO$: 
\begin{align} \label{eq:elbo-general}
{\rm ELBO}^\ell (\theta, \phi) 
    & = \E_{q_\phi(s_{1:N}^\ell)} \bigl[ -\log q_\phi(s_{1:N}^\ell) + \log p_\theta(s_1^\ell) + \textstyle{\sum_{n=2}^N} \log p_\theta(s_n^\ell \,|\, s_{n-1}^\ell) \bigr] \nonumber  \\
    & \quad + \E_{q_\phi(\bm{z}^\ell_{1:N})} \bigl[ -\log q_\phi(\bm{z}^\ell_{1:N}) + \log p_\theta(\bm{z}_1^\ell)  + \textstyle{\sum_{n=1}^N} \log p_\theta (y_n^\ell \,|\, z_n^{\ell,x_n}) \bigr]\nonumber  \\
    & \quad + \E_{q_\phi(\bm{z}^\ell_{1:N})\, q_\phi(s_{1:N}^\ell)} \bigl[ \textstyle{\sum_{n=2}^N} \log p_\theta(\bm{z}_n^\ell \,|\, \bm{z}_{n-1}^\ell, s_n^\ell) \bigr].
\end{align}
The SSM emissions and transitions were introduced in Eqs.~\ref{eq:ssm-traits}-\ref{eq:ssm-observations}, along with the respective initial priors. To allow for a diversity of combinations of learner traits to account for the data, we model the variational posterior across learners,~$q_\phi(s_{1:N})$, as a mixture of Gaussians (see Appendix~\ref{appsec:groupkt-architecture}).

\subsubsection{Inference in continual learning} \label{sec:inference-continual-learning}
In real-world educational settings, a KT model must flexibly adapt its current variational parameters~$\phi_n$ with newly available  interactions~$(x_{n+1}^\ell, t_{n+1}^\ell, y_{n+1}^\ell)$. Retraining on a fixed, augmented history~$\mathcal{H}_{n+1}^\ell$ to obtain an updated~$\phi_{n+1}$ is possible (Eq.~\ref{eq:elbo-general}), but expensive. Instead, in \acro{PSI-KT}, we use the parameters~$\phi_n$ of the current posterior~$q_{\phi_n}(\bm{z}_n^\ell, s_n^\ell)$ to form a next-time prior,
\begin{align}
    \tilde p(\bm{z}_{n+1}^\ell, s_{n+1}^\ell)
    &:= \E_{q_{\phi_n}(\bm{z}_n^\ell, s_n^\ell \,|\, y_{1:n}^\ell)}\big[p_\theta(s_{n+1}^\ell \,|\, s_n^\ell)
    \, p_\theta(\bm{z}_{n+1}^\ell \,|\, s_{n+1}^\ell, \bm{z}_n^\ell)\big].
\end{align}
Due to the Bayesian nature of our model, we can now update this prior with the new evidence~$y_{n+1}^\ell$ at time~$t_{n+1}^\ell$ using \emph{variational continual learning}  \citep[VCL;][]{nguyen2017variational, loo2020generalized}, i.e., by maximizing the ${\rm ELBO}$:
\begin{align} \label{eq:elbo-vcl}
    {\rm ELBO}_{\rm VCL}^\ell(\theta, \phi_{n+1})
    & = \E_{q_{\phi_{n+1}}(s_{n+1}^\ell)} \bigl[- \log q_{\phi_{n+1}}(s_{n+1}^\ell) \bigr]    \nonumber \\[-5pt]  
    & \quad + \E_{q_{\phi_{n+1}}(\bm{z}_{n+1}^\ell)} \bigl[-\log q_{\phi_{n+1}}(\bm{z}_{n+1}^\ell) +           
         \log p_\theta(y_{n+1}^\ell \,|\, z_{n+1}^{\ell,x_{n+1}}) \bigr] \nonumber \\[-5pt]  
    & \quad + \E_{q_{\phi_{n+1}}(\bm{z}_{n+1}^\ell, s_{n+1}^\ell)} \bigl[\log\tilde p(\bm{z}_{n+1}^\ell, s_{n+1}^\ell) \bigr]. 
\end{align}
Maximizing this ${\rm ELBO}_{\rm VCL}^\ell$ allows us to update the parameters~$\phi_{n+1}$ based on a new interaction~$(x_{n+1}^\ell, t_{n+1}^\ell, y_{n+1}^\ell)$ directly from the previous parameters $\phi_n$, i.e., without retraining.

\subsection{Predictions} \label{sec:predictions}
To predict a learner's performance on KC~$x^\ell_{n+1}$ at $t_{n+1}^\ell$, we take the current variational distributions over~$s^\ell_n$ and~$\bm{z}^\ell_n$ and transport them forward by analytically convolving them with the respective transition kernels (Eqs.~\ref{eq:ssm-traits} and~\ref{eq:ssm-states}). We then draw ~$z^{\ell,x_{n+1}}_{n+1}$ from the resulting distribution, and predict the outcome~$\hat{y}_{n+1}^\ell$ by Eq.~\ref{eq:ssm-observations}. %
When predicting multiple steps ahead, we repeat this procedure without conditioning on any of the previously predicted~$\hat{y}_{n+m}^\ell$.

%% file: main_text/4_datasets.tex
\vspace{-5pt}
\section{Evaluations} \label{sec:results}
\input{tables/datasets} %
We argue above that KT for personalized education must predict accurately, scale well with new data, and provide interpretable representations. We now empirically assess these desiderata, comparing \acro{PSI-KT} with up to 8 baseline models across three datasets from online education platforms. Concretely, we evaluate (i) prediction accuracy, quantifying both within-learner prediction and between-learner generalization (Sec.~\ref{sec:exp-prediction}), (ii) scalability in a continual learning setting (Sec.~\ref{sec:exp-continual-learning}), and (iii) interpretability of learner representations and prerequisite relations (Sec.~\ref{sec:exp-interpretability}). 

\textbf{Datasets. } Assistments and Junyi Academy are non-profit online learning platforms for pre-college mathematics.  
 We use Assistments' 2012 and 2017 datasets\footnote{\scriptsize \url{https://sites.google.com/site/assistmentsdata}} (Assist12 and Assist17) and Junyi's 2015 dataset\footnote{\scriptsize \url{https://pslcdatashop.web.cmu.edu/DatasetInfo?datasetId=1198}} \citep[Junyi15;][]{data-junyi}, which in addition to interaction data, provides human-annotated KC relations (see Table~\ref{tab:datasets} and Appendix~\ref{appsec:datasets} for details). 

We select \acro{HLR} from Duolingo and \acro{PPE} as two influential psychologically-informed regression models. From the models that use learnable representations, we include two established deep learning benchmarks, \acro{DKT} and \acro{DKTF}, which capture complex dynamics via LSTM networks, as well as the interpretability-oriented \acro{QIKT}.

%% file: tables/datasets.tex
\begin{wraptable}{r}{5.5cm}
    \vspace{-10pt}
    \caption{Dataset characteristics}
    \label{tab:datasets}
    \vspace{-5pt}
    \small
    \scriptsize
    \begin{tabular}{lccc}
    \toprule
    Dataset $\rightarrow$  & Assist12 & Assist17 & Junyi15\\
    \midrule 
    \# Learners $L$ & 46,674 & 1,709  & 247,606 \\
    \# KCs $K$  & 263 & 102 & 722 \\
    \# Int's / $10^6$ & 3.5 & 0.9 & 26\\
    \bottomrule
    \end{tabular}
    \vspace{-10pt}
\end{wraptable}
 

%% file: main_text/5_1_exp_pred.tex
\subsection{Prediction and generalization performance} \label{sec:exp-prediction} 

\begin{figure}[t]
\vspace{-15pt}
\centering
    {\includegraphics[height=0.8in]{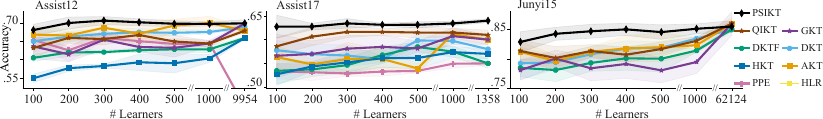}} 
    \vspace{-15pt}
    \caption{Within-learner prediction performance (mean $\pm$\acro{SEM}) as a function of cohort sizes from 100 to the maximum available in each dataset (we omit \acro{HLR} for legibility; see Table~\ref{tab:generalization-results}.)}
    \label{fig:performance-results}
\vspace{-15pt}
\end{figure}

In our evaluations, we mainly focus on prediction and generalization when training on 10 interactions from up to 1000 learners. Good KT performance with little data is key in practical ITS to minimize the number of learners on an experimental treatment \citep[principle of equipoise, similar to medical research;][]{burkholder2021equipoise}, to mitigate the cold-start problem, and to extend the usefulness of the model to classroom-size groups. To provide ITS with a basis for adaptive guidance and long-term learner assessment, we always predict the 10 next interactions.  Figure~\ref{fig:performance-results} shows that \acro{PSI-KT}'s \emph{within-learner prediction performance} is robustly above baselines for all but the largest cohorts ($>$60k learners, Junyi15), where all deep learning models perform similarly.
The advantage of \acro{PSI-KT} comes from its combined modeling of KC prerequisite relations and individual learner traits that evolve in time (see Appendix Fig.~\ref{appfig:ablation-study} for ablations).
The \textit{between-learner generalization} accuracy of the models above, when tested on 100 out-of-sample learners, is shown in Table~\ref{tab:generalization-results}, where fine-tuning indicates that parameters were updated using (10-point) learning histories from the unseen learners. \acro{PSI-KT} shows overall superior generalization except on Junyi15 (when fine-tuning). 

\subsection{Scalability in continual learning} \label{sec:exp-continual-learning}
In addition to training on fixed historical data, we also conduct experiments to demonstrate \acro{PSI-KT}'s scalability when iteratively retraining on additional interaction data from each learner. This parallels real-world educational scenarios, where learners are continuously learning (Sec.~\ref{sec:inference-continual-learning}). 
Each model is initially trained on 10 interactions from 100 learners. We then incrementally provide one data point from each learner, and evaluate the training costs and prediction accuracy. Figure~\ref{fig:continual-learning} shows \acro{PSI-KT} requires the least retraining time, retains the best prediction accuracy, and thus achieves the most favorable cost-accuracy trade-off (details in Appendix~\ref{appsec:continual-learning}).

%% file: main_text/5_2_exp_learner.tex
\subsection{Interpretability of representations}  \label{sec:exp-interpretability}
We now evaluate the interpretability of both learner-specific cognitive traits~$s^\ell$ and the prerequisite graphs $\mathcal{A}$.
We first show that our model captures learner-specific and disentangled traits that correlate with behavior patterns. 
Next, we show that our inferred graphs best align with ground truth graphs, and the edge weights predict causal support on downstream KCs. 

\subsubsection{Learner-specific cognitive traits} \label{sec:exp-interpretability-learner}
For each learner, \acro{PSI-KT} infers four latent traits, each with a clear dynamical role specified by the OU process (Eqs.~\hbox{\ref{eq:ou-process-mean}-\ref{eq:ou-process-marginal}}).
In contrast, high-performance baselines (\acro{AKT}, \acro{DKT}, and \acro{DKTF}) describe learners via 16-dimensional embeddings solely constrained by network architecture and loss minimization. Another model \acro{QIKT} constructs 3-dimensional embeddings with each element connected to scores of knowledge acquisition, knowledge mastery, and problem-solving. We collectively refer to these learner-specific variables as \emph{learner representations}. 
Here, we empirically show that \acro{PSI-KT} representations provide superior interpretability. We ask that learner representations be 1) \emph{specific} to individual learners, 2) \emph{consistent} when trained on partial learning histories, 3) \emph{disentangled} \citep[i.e., component-wise meaningful, as in ][]{bengio2013representation}, and 4) and \emph{operationally interpretable}, so that they can be used to personalize future curricula.
We evaluate desiderata~1-3 with information-theoretic metrics (Table~\ref{tab:interpretability-learner}; see Appendix~\ref{appsec:interpretability-learner} for details), and desideratum~4 with regressions against behavioral outcomes (Table~\ref{tab:interpretability-learner-regression}).

\input{tables/generalization-results}
\begin{figure}[t]
\vspace{-0pt}
\centering
    {\includegraphics[height=1.4in]{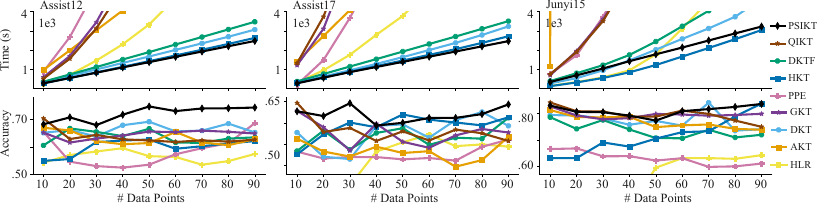}} 
    \caption{Continual learning. \figtop~Cumulative training time. \figbottom~Prediction accuracy on the next 10 time steps. We omit results when time is above, or accuracy is below, the range of the axes.}
    \label{fig:continual-learning}
\vspace{-20pt}
\end{figure}

\input{tables/interpretability-learner}
\textbf{Specificity, consistency, and disentanglement.} Learner representations $s$ should be maximally \emph{specific} about learner identity $\ell$, which can be quantified by the mutual information $\textrm{MI}(s;\ell) = \text{H} (s) - \text{H} (s \,|\, \ell)$ being high, where H denotes (conditional) entropy. 
Additionally, when we infer representations~$s^{\ell_{\rm sub}}$ from different subsets of the interactions of a fixed learner, they should be \emph{consistent}, i.e., each~$s^{\ell_{\rm sub}}$ should be minimally informative about the chosen subset (averaged across subsets), such that $\mathbb{E}_{\ell_\textrm{sub}}  \textrm{MI} (s^{\ell};\ell_\textrm{sub}) = \mathbb{E}_{\ell_\textrm{sub}} \left[ \text{H} (s| \ell)- \text{H} (s| \ell_\textrm{sub}) \right]$ should be low.
Note that sequential subsets are unsuitable for this evaluation, since representations evolve in time to track learners' progression. Instead, we define subsets as groups of KCs whose average presentation time is approximately uniform over the duration of the experiment (see Appendix~\ref{appsec:interpretability-learner-identifiability} for details).
Lastly, learner representations should be \emph{disentangled}, such that each dimension is individually informative about learner identity.
We measure disentanglement with $D_\textrm{KL} (s \| \ell) := \text{H} (s) - \text{H} (s \,|\, \ell)_{\textrm{diag}}$, a form of specificity that ignores correlations across $s^\ell$ dimensions by estimating the conditional entropy only with diagonal covariances.

In empirical evaluations (Table~\ref{tab:interpretability-learner}), \acro{PSI-KT}'s representations offer competitive specificity despite being lower-dimensional, and outperform all baselines in consistency and disentanglement. While disentanglement aids interpretability \citep{freiesleben2022scientific}, it does not itself entail domain-specific meaning for representational dimensions. We now demonstrate that \acro{PSI-KT} representations correspond to clear behavioral patterns, which is crucial for future applications in educational settings. 

\input{tables/interpretability-learner-correlation}
\textbf{Operational interpretability. } \label{exp:operational}
Having shown that \acro{PSI-KT} captures specific, consistent, and disentangled learner features, we now investigate whether these features relate to meaningful aspects of future behavior, which would be useful for scheduling operations for ITS.
We indeed find that the learner representations of \acro{PSI-KT} forecast interpretable behavioral outcomes, such as performance decay or initial performance on novel KCs.
Concretely, consider the observed \emph{one-step performance difference} $\Delta y_n^\ell := y_n^\ell - y_{n-1}^\ell$.
We expect it to be lower for longer intervals $\tau_n^\ell = t_n^\ell - t_{n-1}^\ell$ due to forgetting.
However, we recognize no clear trend when plotting $\Delta y_n^\ell$ over $\tau_n^\ell$ for the Junyi15 dataset (Fig.~\ref{fig:interpretability-learner-regression}, top right).
We can explain this observation because different learners forget on different time scales.
Plotting the same test data instead over scaled intervals $\tau_n^\ell \alpha_n^\ell$ (Fig.~\ref{fig:interpretability-learner-regression}, top center) shows a clear trend against an exponential fit (solid line) with less variability, demonstrating that $\alpha_n^\ell$ (derived from past data only) adjusts for individual learner characteristics and can be interpreted as a personalized rate of forgetting.
Here, the choice of the factor $\alpha_n^\ell$ is motivated by our inductive bias (Eq.~\ref{eq:langevin}).
The trend is much less clear for all baselines:
Fig.~\ref{fig:interpretability-learner-regression} (top left) uses the best fitting component across all learner representations from all baselines (full results in Fig.~\ref{appfig:alpha-regression} in Appendix~\ref{appsec:interpretability-learner-mixed-effect}).
Analogously, when we consider \emph{initial performance on a novel KC}, we find for \acro{PSI-KT} that $\Tilde{\mu}^{\ell,k}_n$ (which aggregates mastery of prerequisites for KC $k$ at time $t_n$, see Eq.~\ref{eq:ou-process-mean}) explains it better than the best baseline Fig.~\ref{fig:interpretability-learner-regression} (bottom panels).
Table~\ref{tab:interpretability-learner-regression} shows that these superior interpretability results are significant and hold across all datasets.
In Appendix~\ref{appsec:interpretability-learner-mixed-effect}, we discuss two more behavioral signatures (performance variability and prerequisite influence) and show they correspond to the remaining components $\gamma_n^\ell$ and~$\sigma_n^\ell$.

%% file: tables/generalization-results.tex
\begin{table}[t]
    \vspace{-15pt}
    \caption{Prediction accuracy. FT indicates additional fine-tuning and $\uparrow$ indicates larger values are better. The \textbf{best model performance} is in bold and the \underline{2nd best} is underlined.
    }
    \vspace{-5pt}
    \label{tab:generalization-results}
    \centering
    \small 
    \scriptsize	
    \begin{tabular}{ll ccccccccc} 
        \toprule
        Dataset & Experiment & \acro{HLR} & \acro{PPE} & \acro{DKT} & \acro{DKTF} & \acro{HKT} & \acro{AKT} & \acro{GKT} & \acro{QIKT} & \acro{PSI-KT} \\ \midrule
        & Within $\uparrow$ & .54$_{.03}$ & .65$_{.01}$ & .65$_{.03}$ & .60$_{.01}$ &  .55$_{.01}$ & \underline{.67}$_{.02}$ & .63$_{.03}$ & .63$_{.03}$ & \textbf{.68}$_{.02}$ \\
        & Between $\uparrow$ & .50$_{.03}$ & .50$_{.02}$ & .55$_{.02}$ & .51$_{.01}$ & .54$_{.00}$  & .58$_{.02}$ & \textbf{.61}$_{.02}$ & \underline{.60}$_{.02}$ & \textbf{.61}$_{.03}$ \\
        \multirow{-3}{*}{Assist12} & w/ FT $\uparrow$ & .52$_{.02}$ & .53$_{.01}$ & .58$_{.00}$ & .55$_{.01}$ & .55$_{.00}$ & \underline{.61}$_{.00}$ & \textbf{.62}$_{.02}$ & .60$_{.03}$ & \textbf{.62}$_{.02}$ \\ \midrule
            
        & Within & .45$_{.01}$ & .53$_{.02}$ & .57$_{.02}$ & .53$_{.03}$ &  .52$_{.03}$ & .56$_{.02}$ & .56$_{.04}$ & \underline{.58}$_{.02}$ & \textbf{.63}$_{.02}$\\
        & Between & .33$_{.03}$ & \underline{.51}$_{.02}$ & .51$_{.00}$ & .48$_{.00}$ & \underline{.51}$_{.02}$ & .47$_{.01}$ & \textbf{.53}$_{.02}$ & .50$_{.02}$ & \textbf{.53}$_{.02}$ \\
        \multirow{-3}{*}{{Assist17}} & w/ FT & .41$_{.04}$ & .51$_{.00}$ & .51$_{.03}$ & .53$_{.01}$ & .51$_{.03}$ & .51$_{.02}$ & \underline{.54}$_{.03}$ & .51$_{.04}$ & \textbf{.56}$_{.02}$ \\ \midrule
        
        & Within & .55$_{.02}$ & .66$_{.03}$ & .79$_{.03}$ & .78$_{.01}$ & .63$_{.02}$ & \underline{.81}$_{.02}$ & .78$_{.02}$ & \underline{.81}$_{.02}$ & \textbf{.83}$_{.02}$ \\
        & Between & .48$_{.02}$ & .55$_{.02}$ & .76$_{.00}$ & .76$_{.02}$ & .61$_{.01}$ & .73$_{.01}$ & \underline{.77}$_{.03}$ & .76$_{.03}$ & \textbf{.79}$_{.03}$ \\
        \multirow{-3}{*}{Junyi15} & w/ FT & .52$_{.00}$ & .65$_{.03}$ & .81$_{.01}$ & \textbf{.84}$_{.01}$ & .64$_{.03}$ & \underline{.83}$_{.00}$ & .79$_{.03}$ & .80$_{.03}$ & .80$_{.02}$ \\ 
        \bottomrule
    \end{tabular}
    \vspace{-10pt}
\end{table}

%% file: tables/interpretability-learner.tex
\begin{wraptable}{br}{6cm}
    \vspace{-0pt}
    \caption{Specificity, consistency, and disentanglement vs. best baseline.} 
    \label{tab:interpretability-learner}
    \vspace{-5pt}
    \small 
    \scriptsize
    \begin{tabular}{llcc} 
    \toprule
    Metric & Dataset & Baseline & \acro{PSI-KT} \\  \midrule
    \multirow{3}{*}{\begin{tabular}[l]{@{}l@{}} Specificity \\ $\textrm{MI}(s;\ell)$ $\uparrow$ \end{tabular}  } 

        & Assist12 & \textbf{8.8} & \underline{8.4} \\
        & Assist17 & \textbf{10.1} & \underline{10.0} \\
        & Junyi15 & \underline{13.5} & \textbf{14.4} \\ \midrule
    \multirow{3}{*}{\begin{tabular}[l]{@{}l@{}} $\text{Consistency}^{-1}$ \\ $\mathbb{E}_{\ell_\textrm{sub}}  \textrm{MI} (s^{\ell};\ell_\textrm{sub})$ $\downarrow$ \end{tabular}} 
        & Assist12 & \underline{12.2} & \textbf{7.4} \\
        & Assist17 & \textbf{6.4} & \textbf{6.4} \\
        & Junyi15 & \underline{7.7}& \textbf{5.0} \\ \midrule
     \multirow{3}{*}{\begin{tabular}[l]{@{}l@{}} Disentanglement \\ $D_\textrm{KL} (s \| \ell)$ $\uparrow$ \end{tabular}} 

         & Assist12 & \underline{2.3} & \textbf{7.4} \\ 
         & Assist17 & \underline{0.6} & \textbf{8.4} \\ 
         & Junyi15 & \underline{5.0} & \textbf{11.5} \\
    \bottomrule
    \end{tabular}
    \vspace{-10pt}
\end{wraptable}

%% file: tables/interpretability-learner-correlation.tex
\begin{table}[b]
    \vspace{-12pt}
      \begin{minipage}[b]{0.50\linewidth}
        \centering
        \includegraphics[width=68mm]{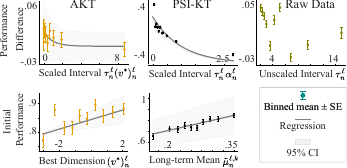}
        \captionof{figure}{Operational interpretability of representations, Junyi15 dataset. See text for axes labels and Appendix~\ref{appsec:interpretability-learner-mixed-effect} for additional results.}
        \label{fig:interpretability-learner-regression}
    \end{minipage}
    \hfill
\begin{varwidth}[b]{0.46\linewidth}
    \centering
    \small
    \scriptsize
    \begin{tabular}{llcc}
    \toprule
        Behavioural \\signature & Dataset & Best Baseline & \acro{PSI-KT} \\ \midrule
        \multirow{3}{*}{\begin{tabular}[c]{@{}l@{}}Performance \\ difference\end{tabular}} 
            & Assist12 & 0.01, .67  & \textbf{0.30}, <.001 \\ %
            & Assist17 & $-$0.03, .30 & \textbf{0.56}, <.001 \\ %
            & Junyi15 & 0.03, .06 & \textbf{0.72}, <.001 \\ \midrule %
        \multirow{3}{*}{\begin{tabular}[c]{@{}l@{}}Initial \\ performance\end{tabular}}  
            & Assist12 & 0.04, .01 & \textbf{0.54}, <.001  \\  %
            & Assist17 & 0.05, .01  & \textbf{3.70}, <.001  \\ %
            & Junyi15 & 0.04, .02 & \textbf{0.92}, <.001 \\  %
     \bottomrule
     & & & 
    \end{tabular}
    \caption{Coefficients and $p$-values of regressions  relating $\exp(-\alpha_n^\ell\,\tau_n^\ell)$ and $\Tilde{\mu}_n^{\ell,k}$ to unseen behavioral data across datasets.} %
    \label{tab:interpretability-learner-regression}
  \end{varwidth}
\vspace{-15pt}
\end{table}

%% file: main_text/5_3_exp_graph.tex
\subsubsection{Prerequisite graph} \label{sec:exp-interpretability-graph}
\begin{figure}[t]
\vspace{-0pt}
\centering
    {\includegraphics[height=2.2in]
    {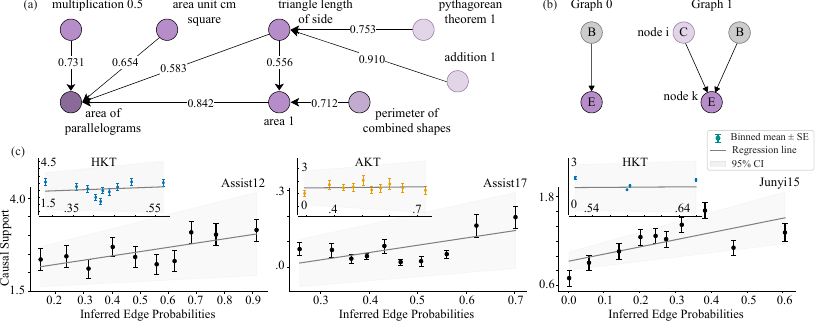}
    } 
    \vspace{-10pt}
    \caption{Graph interpretability.
    \captiona~Subgraph inferred by \acro{PSI-KT} on the Junyi15 dataset, showing prerequisites of target KC `area of parallelograms`.  
    \captionb~Hypothesized causal graphs, where Graph 1 assumes a causal relationship exists from KC~$i$ to KC~$k$, while Graph 0 is the null hypothesis. 
    \captionc~Regression of edge probabilities against causal supports. Insets show the best baseline model.}
    \label{fig:graph-visualization}
\vspace{-18pt}
\end{figure}

\acro{PSI-KT} infers a prerequisite graph based on all learners' data, which helps it to generalize to unseen learners. Beyond helping prediction, reliable prerequisite relations are an essential input for curriculum design, motivating our interest in their interpretability. Figure~\ref{fig:graph-visualization}a shows an exemplary inferred subgraph with the prerequisites of a single KC.
To quantitatively evaluate the graph, we (i) measure the alignment of the inferred vs. ground-truth graphs 
and (ii) correlate inferred prerequisite probability with a Bayesian measure of causal support obtained from unseen behavioral data.

\textbf{Alignment with ground-truth graphs. }
We analyze the Junyi15 dataset, which uniquely provides human-annotated evaluations of prerequisite and similarity relations between KCs. We discuss here the alignment of prerequisites and leave similarity for Appendix~\ref{appsec:graph}. The Junyi15 dataset provides both an expert-identified prerequisite for each KC,and crowd-sourced ratings (6.6 ratings on average on a 1-9 scale). To compare with expert annotations, we compute the rank of each expert-identified prerequisite relation $i \rightarrow k$ in the relevant sorted list of inferred probabilities $\{a^{jk}\}_{j=1}^K$ and take the harmonic average \cite[mean reciprocal rank, MRR;][]{yang2014embedding}.  Next, we compute the negative log-likelihood (nLL) of inferred edges~$a^{ik}$ using a Gaussian estimate of the (rescaled) crowd-sourced ratings for the $i \rightarrow k$ KC pair. We finally calculate the Jaccard similarity (JS) between the set of inferred edges ($a^{ik}>0.5$) and those identified by experts as well as crowd-sourced edges with average ratings above 5.
The results in Table~\ref{tab:graph-comparison-results} (left columns) consistently highlight \acro{PSI-KT}'s superior performance across all criteria (see Appendix~\ref{appsec:graph-metric-detail} for details).

\input{tables/graph-comparison-results}

\textbf{Causal support across consecutive interactions. }
For education applications, we are interested in how KC dependencies impact learning effectiveness. If KC~$i$ is a prerequisite of KC~$k$, mastering KC~$i$ contributes to mastering KC~$k$, indicating a causal connection. 
In this analysis, we show that inferred edge probabilities~$a^{ij}$ (Eq.~\ref{eq:adjacency-matrix}) correspond to causal $\text{support}_{i \rightarrow k}$ (Eq.~\ref{eq:causal-support}), derived from behavioral data through Bayesian causal induction \citep{griffiths2009theory}. 
Specifically, we model the relationship between a candidate cause~$C$ and effect~$E$, i.e., a pair of KCs in our case, while accounting for a constant background cause~$B$, representing the learner's overall ability and the influences of other KCs.
We consider two hypothetical causal graphs, where Graph 0~$G_{i \nrightarrow k}$ represents the null hypothesis of no causal relationship, and Graph 1~$G_{i \rightarrow k}$ assumes the causal relationship exists, i.e. correct performance on KC~$i$ causally supports correct performance on KC~$k$ (Fig.~\ref{fig:graph-visualization}b). 
We estimate causal support for each pair of KCs~$i \rightarrow k$ based on all consecutive interactions in the behavioral data~$\mathcal{H}$ from KC~$i$ at time~$t_n$ to KC~$k$ at time~$t_{n+1}$, as a function of the difference in log-likelihoods of the two causal graphs (see Appendix~\ref{appsec:causal-support} for details):
\begin{align} \label{eq:causal-support}
    \text{support}_{i \rightarrow k} :=\log P(\mathcal{H} \,|\, G_{i \rightarrow k})-\log P (\mathcal{H} \,|\, G_{i \nrightarrow k}).
\end{align} 
We then use regression to predict $\text{support}_{i \rightarrow k}$ as a function of edge probabilities~$a^{ij}$ inferred from different models. The results are visualized in Figure~\ref{fig:graph-visualization}c and summarized in Table~\ref{tab:graph-comparison-results} (right). The larger coefficients indicate that our inferred graphs possess superior operational interpretability (Sec.~\ref{sec:exp-interpretability}).

%% file: tables/graph-comparison-results.tex
\begin{table}[t]
    \centering
    \vspace{-15pt}
    \caption{
        \figleft~Alignment of inferred graphs with annotated graphs for the Junyi15 dataset.
        \figright~Regression coefficients and $p$-values relating causal support to inferred edge probabilities.
        All baseline models either lack significance or negatively predict causal support~(Appendix~Fig.~\ref{appfig:graph-causal-support-regression}).}
    \label{tab:graph-comparison-results}
    \vspace{-5pt}
    \small
    \scriptsize
    \begin{tabular}{lcccccccc}
    \toprule  
        Metric & \multicolumn{1}{c}{MRR $\uparrow$} & \multicolumn{1}{c}{JS expert $\uparrow$} & \multicolumn{1}{c}{JS crowd $\uparrow$} & \multicolumn{1}{c}{nLL $\downarrow$} && \multicolumn{3}{c}{ coefficient $\uparrow$, $p$-value $\downarrow$} \\ \cmidrule{2-5}  \cmidrule{7-9} 
        Dataset & \multicolumn{4}{c}{Junyi15} && Assist12 & Assist17 & Junyi15  \\ \midrule        
        Best Baseline & .0082 & .0015 & .0047 & \textbf{3.03} && 1.05, .253 & 0.22, .792 &  0.42, .593 \\
        \acro{PSI-KT} & \textbf{.0086} & \textbf{.0019} & \textbf{.0095} & 4.11 && \textbf{1.15}, .003 & \textbf{0.28}, <.001 &  \textbf{0.97}, <.001 \\     
    \bottomrule
    \end{tabular}
    \vspace{-0pt}
\end{table}

%% file: main_text/6_discussion.tex
\section{Discussion} \label{sec:discussion}
We propose \acro{PSI-KT} as a novel approach to knowledge tracing (KT) with compelling properties for intelligent tutoring systems: superior \textbf{p}redictive accuracy, excellent continual-learning \textbf{s}calability, and \textbf{i}nterpretable representations of learner traits and prerequisite relationships. 
We further find that \acro{PSI-KT} has remarkable predictive performance when trained on small cohorts whereas baselines require training data from at least 60k learners to reach similar performance. An open question for future KT research is how to combine \acro{PSI-KT}'s unique continual learning and interpretability properties with performance that grows beyond this extreme regime.
We use an analytically marginalizable Ornstein-Uhlenbeck process for knowledge states in \acro{PSI-KT}, resulting in an exponential forgetting law, similar to most recent KT literature. Future work should support ongoing debates in cognition by offering alternative modeling choices for memory decay \cite[e.g., power-law;][]{wixted1997genuine}, thus facilitating empirical studies at scale. 
And while our model already normalizes reciprocal dependencies in the prerequisite graph, we anticipate that enforcing regional or global structural constraints, such as acyclicity, may benefit inference and interpretability.
Although we designed \acro{PSI-KT} with general structured domains in mind, our empirical evaluations were limited to mathematics learning by dataset availability. We highlight the need for more diverse datasets for structured KT research to strengthen representativeness in ecologically valid contexts.
Overall, our work combines machine learning techniques with insights from cognitive science to derive a predictive and scalable model with psychologically interpretable representations, thus laying the foundations for personalized and adaptive tutoring systems.

%% file: appendix/7_1_elbo.tex
\section{Appendix}
The Appendix is organized as follows: 
\begin{itemize}[noitemsep,nolistsep]
    \item Appendix~\ref{appsec:elbo} and \ref{appsec:continual-elbo} provides a detailed derivation of our objective function $\rm ELBO$ in two scenarios: inference involving complete learning histories (Sec.~\ref{sec:inference-entire-history}) and inference for in a continual learning setting (Sec.~\ref{sec:inference-continual-learning}).
    \item Appendix~\ref{appsec:baseline-model} provides in-depth descriptions of baseline models, and the details and the selection criterion of the three datasets we use for experiments. 
    \item Appendix~\ref{appsec:groupkt-architecture} describes the \acro{PSI-KT} architecture in full detail, including its hyperparameters.
    \item Appendix~\ref{appsec:exp-prediction} provides additional prediction results. We show the average accuracy, average f1-score, and their standard error of Fig.~\ref{fig:performance-results} in the prediction experiment given entire learning histories. We also show the average accuracy score and their standard error of Fig.~\ref{fig:continual-learning} in the prediction experiment for continual learning setup.  
    \item Appendix~\ref{appsec:interpretability-learner} describes the details of the experiment setup and how we derive the metrics for specificity, consistency, and disentanglement. We also provide comprehensive results on the operational interpretability of baseline models. 
    \item Appendix~\ref{appsec:graph} elaborates on the graph assessment framework, including details about the alignment metrics and a discussion of causal support, and extends the main text evaluations to all datasets. 
\end{itemize}

\subsection{ELBO of the hierarchical SSM} \label{appsec:elbo}
In this section, we derive the single-learner ${\rm ELBO}$, Eqs.~\ref{eq:elbo-general}-~\ref{eq:elbo-vcl} in the main text. For clarity, we omit the superindex~$\ell$ in these derivations. Note that the parameters~$\phi$ and~$\theta$ are global, i.e., they are optimized based on the entire interaction data across learners.

In variational inference (VI), we approximate an intractable posterior distribution~$p_\theta(z \,|\, y)$ with $q_\phi(z \,|\, y)$ from a tractable distribution family. We learn~$\phi$ and~$\theta$ together by maximizing the evidence lower bound ($\rm ELBO$) of the marginal likelihood \citep{blei2017variational, attias1999variational}, given by~$\log p_\theta(y) \geq {\rm ELBO}(\theta, \phi) = \E_{q_\phi(z \,|\, y)}\left[ -\log q_\phi(z \,|\, y)+\log p_\theta(y, z) \right]$. 
The two terms in the~$\rm ELBO$ represent the entropy of the variational posterior distribution,~$\text{H}\bigl[ q_\phi(z \,|\, y) \bigr] = \E_{q_\phi(z | y)} \left[ -\log q_\phi(z \,|\, y) \right]$, and the log-likelihood of the joint distribution of observations and latent states~$\E_{q_\phi(z|y)}\log p_\theta(y, z)$.

We now formulate the~$\rm ELBO$ for our hierarchical SSM (see Fig.~\ref{fig:model}) with two layers of latent states. 
We assume that fixed learning histories~$\mathcal{H}_{1:N}$ until time~$t_N$ are available and we use capital~$N$ to represent the fixed time point. We approximate the posterior~$p_\theta(\bm{z}_{1:N}, s_{1:N} \,|\, y_{1:N})$ using the mean-field factorization,~$q_\phi(\bm{z}_{1:N}, s_{1:N} \,|\, y_{1:N}) = q_\phi(\bm{z}_{1:N})\, q_\phi(s_{1:N})$:
\begin{align} \label{appeq:hssm-elbo}
    {\rm ELBO} (\theta, \phi) 
    & = \text{H} \bigl[ q_\phi(\bm{z}_{1:N}, s_{1:N} \,|\, y_{1:N}) \bigr] + \E_{q_\phi(\bm{z}_{1:N}, s_{1:N} | y_{1:N})} \log p_\theta(y_{1:N}, \bm{z}_{1:N}, s_{1:N}) \nonumber \\
    & = \E_{q_\phi(\bm{z}_{1:N}, s_{1:N} \,|\, y_{1:N})} \bigl[ 
        - \log q_\phi(\bm{z}_{1:N}, s_{1:N} \,|\, y_{1:N}) + 
        \log p_\theta(y_{1:N}, \bm{z}_{1:N}, s_{1:N})  \bigr] \nonumber \\
    & = \E_{q_\phi(\bm{z}_{1:N})\, q_\phi(s_{1:N})} \bigl[ -\log q_\phi(\bm{z}_{1:N}) - \log q_\phi(s_{1:N}) + \log p_\theta(y_{1:N}, \bm{z}_{1:N}, s_{1:N}) \bigr].
\end{align}

In the generative model of \acro{PSI-KT}, the observation~$y_n$ at time~$t_n$ depends on the particular knowledge state~$z_n^k$ associated with the interacted KC~$k=x_n$. All knowledge states~$z_n$ rely on previous states~$z_{n-1}$ and cognitive traits~$s_n$, which themselves are influenced by~$s_{n-1}$. 
Thus, we can factorize the joint distribution~$p_\theta(y_{1:N}, \bm{z}_{1:N}, s_{1:N})$ in Eq.~\ref{appeq:hssm-elbo} over all latent states and observations: 
\begin{align} \label{appeq:joint-dist-yzs}
    p_\theta (y_{1:N}, \bm{z}_{1:N}, s_{1:N}) 
    & = p_\theta(s_{1:N}) \, p_\theta(\bm{z}_{1:N} \,|\, s_{1:N})\, \prod_{n=1}^N p_\theta(y_n \,|\, z_n^{x_n}) \nonumber \\
    & = p_\theta(s_1)\, p_\theta(\bm{z}_1)\, \prod_{n=2}^N p_\theta(s_n \,|\, s_{n-1})\, p_\theta(\bm{z}_n \,|\, \bm{z}_{n-1}, s_n)\, \prod_{n=1}^N p_\theta (y_n \,|\, z_n^{x_n}).
\end{align}
Here, $p_\theta(s_1)$ and $p_\theta(\bm{z}_1)$ are the Gaussian initial priors for the latent states.

By incorporating the factorized joint distribution (Eq.~\ref{appeq:joint-dist-yzs}), the~$\rm ELBO$ for \acro{PSI-KT} can be derived as follows:
\begin{align}
    {\rm ELBO} (\theta, \phi) 
    & = \E_{q_\phi(\bm{z}_{1:N})\, q_\phi(s_{1:N})} \bigl[ -\log q_\phi(\bm{z}_{1:N}) 
        - \log q_\phi(s_{1:N}) + \log p_\theta(y_{1:N}, \bm{z}_{1:N}, s_{1:N}) \bigr] \nonumber \\
    & = \E_{q_\phi(\bm{z}_{1:N})\, q_\phi(s_{1:N})}
        \Bigl[ 
            -\log q_\phi(\bm{z}_{1:N}) - \log q_\phi(s_{1:N}) + \log p_\theta(s_1) + \log p_\theta(\bm{z}_1)  \nonumber \\
           & \phantom{=\E_{q_\phi(\bm{z}_{1:N})\, q_\phi(s_{1:N})}} + \sum_{n=2}^N \log p_\theta(s_n \,|\, s_{n-1}) + \sum_{n=2}^N \log p_\theta(\bm{z}_n \,|\, \bm{z}_{n-1}, s_n) \nonumber \\
           & \phantom{=\E_{q_\phi(\bm{z}_{1:N})\, q_\phi(s_{1:N})}} + \sum_{n=1}^N \log p_\theta (y_n \,|\, z_n^{x_n})
       \Bigr] \nonumber \\ 
    & = \E_{q_\phi(s_{1:N})} \bigl[ -\log q_\phi(s_{1:N}) + \log p_\theta(s_1) 
        + \sum_{n=2}^N \log p_\theta(s_n \,|\, s_{n-1}) \bigr] \nonumber \\
        & \quad + \E_{q_\phi(\bm{z}_{1:N})} \bigl[ -\log q_\phi(\bm{z}_{1:N}) + \log p_\theta(\bm{z}_1) + \sum_{n=1}^N \log p_\theta (y_n \,|\, z_n^{x_n}) \bigr] \nonumber \\
        & \quad + \E_{q_\phi(\bm{z}_{1:N})\, q_\phi(s_{1:N})} \bigl[ \,\sum_2^n \log p_\theta(\bm{z}_n \,|\, \bm{z}_{n-1}, s_n) \bigr].
\end{align}

\subsection{Extension to continual learning} \label{appsec:continual-elbo}
We now extend the $\rm ELBO$ to the continual learning setting, where we observe learning performances~$y_{1:n}$ sequentially. Here we use the lower case~$n$ to indicate the running time index.
We seek the posterior distribution~$p_\theta(\bm{z}_n, s_n \,|\, y_{1:n})$ at each interaction time~$t_n$ given all observations so far. Usually, one would approximate the posterior with the variational posterior distribution~$q_{\phi_n}(\bm{z}_n, s_n \,|\, y_{1:n}) = q_{\phi_n}(\bm{z}_n) q_{\phi_n}(s_n)$ using the mean-field factorization (Eq.~\ref{appeq:hssm-elbo}). In that case, the inference process consists of maximizing the ${\rm ELBO}(\theta, \phi_n)$ only over ~$\phi_n$:
\begin{align}
    {\rm ELBO}(\theta, \phi_n)=\E_{q_{\phi_n}(\bm{z}_n) q_{\phi_n}(s_n)}[-\log q_{\phi_n}(\bm{z}_n) - \log q_{\phi_n}(s_n) + \log p_\theta(y_{1:n}, \bm{z}_n, s_n)].
\end{align}

However, it is challenging to calculate the joint distribution~$p_\theta(y_{1:n}, \bm{z}_n, s_n)$ in our setup, since it requires marginalizing the full joint distribution~$p_\theta(y_{1:n}, \bm{z}_{1:n}, s_{1:n})$ over all~$\bm{z}_{n^\prime}$ and $s_{n^\prime}$ with~$n^\prime<n$. 

Henceforth, we aim to reconfigure the objective function~${\rm ELBO}(\theta, \phi_n)$, which involves the variational posterior distribution~$q_{\phi_n}(\bm{z}_n, s_n \,|\, y_{1:n})$ at the time point~$t_n$, to establish a linkage with the posterior~$q_{\phi_{n-1}}(\bm{z}_{n-1}, s_{n-1} \,|\, y_{1:n-1})$ observed at the preceding time point~$t_{n-1}$. 
By doing so, we can recursively optimize the variational parameters~$\phi_n$ whenever new observations~$y_n$ are received \citep{nguyen2017variational}, wherein the initialization draws from the parameters~$\phi_{n-1}$ obtained at time~$t_{n-1}$. 

First, we show that the marginal joint distribution~$p_\theta(y_{1:n}, \bm{z}_n, s_n)$ is proportional to the prior distribution~$p_\theta(\bm{z}_n, s_n \,|\, y_{1:n-1})$ on $s_n, \bm{z}_n$ at~$t_{n-1}$: 
\begin{align} \label{appeq:joint-vcl}
    p_\theta(y_{1:n}, \bm{z}_n, s_n) \propto p_\theta(\bm{z}_n, s_n \,|\, y_{1:n-1})\, p_\theta(y_n \,|\, \bm{z}_n).
\end{align}

Second, we show how the prior distribution~$p_\theta(\bm{z}_n, s_n \,|\, y_{1:n-1})$ can be formulated using the posterior~$q_{\phi_{n-1}}(\bm{z}_{n-1}, s_{n-1} \,|\, y_{1:n-1})$ at the previous time~$t_{n-1}$, which evolves for a single time step: 
\begin{align}
    p_\theta(\bm{z}_n, s_n \,|\, y_{1:n-1})
    & = \int p_\theta(\bm{z}_n, \bm{z}_{n-1}, s_n, s_{n-1} \,|\, y_{1:n-1}) {\rm d} s_{n-1}\,{\rm d} \bm{z}_{n-1} \nonumber \\
    & =\int p_\theta(\bm{z}_{n-1}, s_{n-1} \,|\, y_{1:n-1})\, p_\theta(s_n \,|\, s_{n-1})\, p_\theta(\bm{z}_n \,|\, s_n, \bm{z}_{n-1})\ {\rm d} s_{n-1}\,{\rm d} \bm{z}_{n-1} \nonumber \\
    & = \int q_{\phi_{n-1}}(\bm{z}_{n-1}, s_{n-1} \,|\, y_{1:n-1})\, p_\theta(s_n \,|\, s_{n-1})\, p_\theta(\bm{z}_n \,|\, s_n, \bm{z}_{n-1})\ {\rm d} s_{n-1}\,{\rm d} \bm{z}_{n-1} \nonumber \\
    & = \int q_{\phi_{n-1}}(\bm{z}_{n-1})\, q_{\phi_{n-1}}(s_{n-1})\, p_\theta(s_n \,|\, s_{n-1})\, p_\theta(\bm{z}_n | s_n, \bm{z}_{n-1}) {\rm d}s_{n-1} {\rm d} \bm{z}_{n-1} \nonumber \\
    & = \underbrace{\E_{q_{\phi_{n-1}}(s_{n-1})} \bigl[ p_\theta(s_n \,|\, s_{n-1}) \bigr]}_{:=\tilde{p}_{\phi_{n-1}}(s_n)} \underbrace{\E_{q_{\phi_{n-1}}(\bm{z}_{n-1})} \bigl[ p_\theta(\bm{z}_n \,|\, s_n, \bm{z}_{n-1}) \bigr]}_{:=\tilde{p}_{\phi_{n-1}}(\bm{z}_n \,|\, s_n)}.
\end{align}

Substituting $p_\theta(s_n, \bm{z}_n \,|\, y_{1:n-1}) = \tilde{p}_{\phi_{n-1}}(s_n)\, \tilde{p}_{\phi_{n-1}}(\bm{z}_n \,|\, s_n)$ using our variational approximation in Eq.~\ref{appeq:joint-vcl} in the {\rm ELBO}, we finally arrive at the objective function for variational continuous learning:
\begin{align} \label{appeq:elbo-vcl-detail}
    {\rm ELBO}_{\rm VCL}(\theta, \phi_n)
    & = \E_{q_{\phi_n}(s_n)\, q_{\phi_n}(\bm{z}_n)} \big[-\log q_{\phi_n}(s_n) -\log q_{\phi_n}(\bm{z}_n) + \log \tilde{p}_{\phi_{n-1}}(s_n) \nonumber \\
    & \phantom{ = \E_{q_{\phi_n}(s_n)\, q_{\phi_n}(\bm{z}_n)}} + \log \tilde{p}_{\phi_{n-1}}(\bm{z}_n | s_n) + \log p_\theta(y_n \,|\, \bm{z}_n^{x_n})\big]  \nonumber \\ 
    & = \E_{q_{\phi_n}(s_n)} \bigl[ - \log q_{\phi_n}(s_n) + \log \tilde{p}_{\phi_{n-1}}(s_n) \bigr]  \nonumber \\
    & \quad + \E_{q_{\phi_n}(\bm{z}_n)} \bigl[ - \log q_{\phi_n}(\bm{z}_n) + \log p_\theta(y_n \,|\, z_n^{x_n}) \bigr] \nonumber\\ 
    & \quad + \E_{q_{\phi_n}(\bm{z}_n)\, q_{\phi_n}(s_n)} \bigl[ \log \tilde{p}_{\phi_{n-1}}(\bm{z}_n \,|\, s_n) \bigr].
\end{align}
This provides a derivation of Eq.~\ref{eq:elbo-vcl} as presented in the main text. Here our focus lies in the optimization of the parameters~$\phi_n$, while holding constant the parameters~$\phi_{n-1}$ acquired from the preceding time step.

%% file: appendix/7_2_baseline.tex
\subsection{Baseline models and datasets} \label{appsec:baseline-model}
\subsubsection{Baselines} \label{appsec:baseline}
KT models aim to predict the performance~$\hat{y}_n^\ell$ of the presented KC~$x_n^\ell$ at time ~$t_n^\ell$ for each learner~$\ell$, which amounts to learning the mapping $\hat{y}_n^\ell = f_\theta(\mathcal{H}_{n^\prime<n}^\ell)$ (Sec.~\ref{sec:background-kt}). 
Because baseline models lack learner-specific parameters, we here describe the prediction process for a single learner, and omit the superindex~$\ell$ for clarity. Extending to multiple learners is straightforward since all parameters~$\theta$ are global. 
We use~$\tau_n := t_n - t_{n-1}$ to represent the time interval between consecutive interactions of a learner~$\ell$, and the KC-specific interval~$\tau_n^k := t_n^k - t_{n-1}^k$ for consecutive interactions with the same KC~$k$. The number of practice repetitions for each KC~$k$ up to time~$t_n$ is denoted as $c_n^k$. The dimension of embeddings~$D$ equals 16 in our experiments. 
\input{tables/models}

We compare with eight baseline models (Sec.~\ref{sec:results}):

\begin{enumerate}[label=\alph*)]
    \item \acro{HLR} \citep{kt-hlr}  uses the cumulative counts of correct, incorrect, and total interactions of KC~$k$ until time~$t_n$, collectively denoted $\bm{c}_n^k = {\begin{bmatrix} c_n^{k,1} c_n^{k, 0} c_n^k \end{bmatrix}}^\intercal \in\mathbb{R}^3$, as well as the last interval~$\tau_n^k$. 
    When a learner interacts with KC~$x_n$ at time~$t_n$, \acro{HLR} predicts the probability of a correct performance as:
    \begin{align}
        \hat{y}_n := 2^{ -\tau_n^k / h_n^k}, 
        \quad \text{with memory half-life }\, h_n^k :=2^{\theta^\intercal \bm{c}_n^k} \; \text{and}\; k=x_n.
    \end{align} 
    The learnable weights~$\theta \in \mathbb{R}^3$ modulate the influences of correct, incorrect, and total interaction counts. The training process of \acro{HLR} does not differentiate features from different KCs or learners, thus \acro{HLR} cannot model the relational structure of KCs or any learner-specific characteristics. 
    
    \item \acro{PPE} \citep{kt-ppe} is similar to \acro{HLR} in predicting performance as a function of interaction histories. It defines the activation~$m_n^k$ of KC~$k$ at time~$t_n$, $m_n^k:= {(c_n^k)}^\beta\,(T_n^k)^{-\alpha}$ with separate forgetting rate~$\alpha$ and learning rate~$\beta$. The forgetting term~$T_n^k$ is a function of the interaction history, which it summarizes as a weighted average of times~$\tau_n^k$ elapsed between the exposures to a given KC prior to~$t_n$:
    \begin{align}
        T_n^k :=\sum_{i=1}^{n-1}\, w_i^k\, \tau_i^k, 
        \quad \text{with } w_i^k =(\tau_i^k)^{-\eta} \sum_{j=1}^{n-1}\, {(\tau_j^k)}^{\eta}.
    \end{align}
    The forgetting rate~$\alpha$ is a function of a \emph{stability term}~$\kappa$ and a cumulative average of interval durations between KC exposures modulated by the slope~$\lambda$:
    \begin{align}
        \alpha_n^k = \kappa + \lambda \Bigg(\frac{1}{n-1} \sum_{j-1}^{n-1} \frac{1}{\ln (\tau_j^k+e)} \Biggr).
    \end{align}
    Finally, \acro{PPE} treats performance~$\hat{y}_n$ as a logistic function of~$m_n^k$ with $k=x_n$. The learnable parameters are $\theta = \{\beta, \eta, \kappa, \lambda \}$. 
    
    \item \acro{DKT} \citep{kt-dkt} infers two separate embeddings~$u^k = \{u^{k,0}, u^{k,1}\}$ for each KC~$k$, depending on performance. Here, $u^{k,0}, u^{k,1} \in \mathbb{R}^D$ represent incorrect interactions and correct interactions on KC~$k$, respectively, and are shared across all learners, with $D$ being the dimensionality of the embeddings. 
    \acro{DKT} trains an LSTM \citep{lstm} over $\mathcal{H}_{n^\prime<n}$ to encode the combined information of KC indices and performance. 
    For each learner~$\ell$ and all time points~$t_{n^\prime<n}$, \acro{DKT} takes performance embeddings~$u^{x_{n^\prime},0}$ of interacted KC~$x_{n^\prime}$ as the input when the performance~$y_{n^\prime}$ is incorrect, or $u^{x_{n^\prime},1}$ for correct performance, i.e., \acro{DKT} takes inputs~$\bm{u}_{n^\prime} = u^{x_{n^\prime}, y_{n^\prime}}$ for all~$t_{n^\prime}$ with $n^\prime<n$. \acro{DKT} then predicts the subsequent performance on all KCs~$ \hat{\bm{y}}_n =  [y_n^1, \ldots,  y_n^K]^\intercal \in \mathbb{R}^K$, and chooses only the interacted one, i.e., the $x_n$-th dimension: 
    \begin{align}
        & \bm{h}_n = \text{LSTM}(\bm{u}_{n^\prime<n}; \bm{W}_h, \bm{b}_h) \nonumber \\ 
        & \bm{\hat{y}}_n = \sigma(\bm{W}_{\hat{y} h} \bm{h}_n + b_{\hat{y}}) \nonumber \\
        & \hat{y}_n = \bm{\hat{y}}_n[x_n].
    \end{align}
    Thus, $\theta$ for \acro{DKT} consists of the neural network parameters~$\bm{W}_h, \bm{b}_h, \bm{W}_{\hat{y} h}, \bm{b}_{\hat{y}}$. 
    
    \item \acro{DKTF} \citep{kt-dktf} uses the same LSTM architecture and the same combined KC-performance embeddings, $u^k = \{u^{k,0}, u^{k,1}\}$ that we described above for \acro{DKT}. 
    \acro{DKTF} uses additional 3-dimensional features~$\bm{t}_n := [\tau_n, \tau_n^k, c_n^k]$ representing the KC-unspecific and KC-specific intervals defined above and the cumulative interaction counts~$c_n^k$ for KC~$k$ until time~$t_n$. 
    Then, for inputs of every time point~$t_{n^\prime}$, \acro{DKTF} concatenates the time information~$\bm{t}_{n^\prime}$ with KC performance inputs $\bm{u}_{n^\prime}$ for the interacted KC~$x_{n^\prime}$. 
    \acro{DKTF} predicts future performances following the same architecture based on concatenated input~$[\bm{t}_{n^\prime<n}; \bm{u}_{n^\prime<n}]$. 
    
    \item \acro{HKT} \citep{kt-hkt} is the most similar model to our \acro{PSI-KT}. It uses a Hawkes process to model the structural influence on the state of KC~$k$ due to every other KCs state in the past interactions~$i \in x_{n^\prime<n}$ until time~$t_n$:
    \begin{align}
        & m_n^k=\lambda_k + \sum_{i \in x_{n^\prime<n}} a^{i,k}_n\, \kappa(t^k_n-t^i_{n^\prime}) \nonumber \\
        & \kappa(t^k_n-t^i_{n^\prime}) = \exp \Bigl( -\bigl(1+\beta^{i,k}_n \log(t^k_n-t^i_{n^\prime}) \bigr) \Bigr).
    \end{align}
    Here $m_n^k$ includes a base level~$\lambda_k$ and all previous learned KCs' influences~$a^{i,k}_n$ weighted by the a temporal exponential decay~$\kappa(t^k_n-t^i_{n^\prime})$. 
    The base level~$\lambda_k$ reflects aspects of KC~$k$ but also of the specific assignments that were interacted with at time point~$t_n$, given that distinct assignments can provide practice for a single KC. %
    To model cross-KC influences~$a^{i,k}_n$, \acro{HKT} infers embeddings~$\{u^{k,0}_a, u^{k,1}_a, u^k_a\} \in \mathbb{R}^D$ for each KC~$k$. Here $u^{k,0}_a, u^{k,1}_a$ are defined similarly to the \acro{DKT} embeddings, whereas $u^k_a$ only depends on KC identity~$k$. When interacting with KC~$k$ at time~$t_n$, the influence on its state due to having interacted with KC~$i$ at time~$t_{n^\prime}$ with performance~$y_{n^\prime}$ is estimated as $ a^{i,k}_n = (u^{i,y_{n^\prime}}_a)^\intercal u^k_a $. For the coefficient~$\beta^{i,k}_n$, \acro{HKT} estimates three additional KC-specific embeddings $\{u^{k,0}_\beta, u^{k,1}_\beta, u^k_\beta\}$, and follows similar calculations as for $a$ above. \acro{HKT} also predicts the performance~$\hat{y}_n$ as a logistic function of $m_n^k$ with~$k=x_n$. 
    
    \item \acro{AKT} \citep{kt-akt} is a transformer-based model \citep{transformer} that learns the structure of KCs implicitly from the self-attention weights. Unlike LSTM models, which only capture temporal information, \acro{AKT} captures both temporal and structural relations.
    Specifically, \acro{AKT} first initializes three embeddings~$\{u^{k,0}_a, u^{k,1}_a, u^k_a\}$ for each KC~$k$ and a scalar~$\mu^q$ for each specific assignment~$q$, representing its difficulty level. For each KC~$k$, these embeddings are defined as in \acro{HKT} to separately reflect KC-specific correct/incorrect interactions and KC identity. 
    However, \acro{AKT} combines these representations with three additional embeddings~$\{u^{k,0}_b, u^{k,1}_b, u^k_b\}$, in order to account for difficulty levels. When a learner interacts at time~$t_{n^\prime}$ with an assignment~$q$ related to KC~$k$, the KC identity embedding becomes~$u^k = u^k_a + \mu^q u^k_b$; after assessing the performance at time~$t_{n^\prime}$, the interaction embeddings are similarly updated as $u^{k,y_{n^\prime}} = u^{k,y_{n^\prime}}_b + \mu^q u^{k,y_{n^\prime}}_b$. 
    Consequently, a learner's entire interaction history~$\mathcal{H}_{n^\prime<n}$ is represented as a sequence of these combined KC-interaction-difficulty embeddings. \acro{AKT} processes these sequential embeddings as input, using KC embeddings as queries and keys, and interaction embeddings as values within its attention mechanism.
    To predict performance~$\hat{y}_n$ given the KC and assignment, \acro{AKT} uses the KC embeddings at time~$t_n$ to compare with previous queries and keys in the learning history, and then extract the value. Details about the transformer architecture can be found in \cite{kt-akt}. 

    \item \acro{GKT} \citep{kt-gkt} applies a graph neural network to leverage the graph-structured nature of knowledge.  Like \acro{AKT}, \acro{GKT} initializes three embeddings~$\{u^{k,0}, u^{k,1}, u^k \}$ for each KC~$k$, but instead of only using the embeddings to determine the KC relations, \acro{GKT} learns an additional undirected KC graph, represented by its adjacency matrix $\bm{A}$. Here $a_{ij} = 1$ represents KC~$i$ and KC~$j$ are related, i.e., there is information transmission among KC~$i$ and KC~$j$ every time the model gets updated. To use the KC relations, \acro{GKT} first aggregates the hidden states~$\bm{h}^k_n$ and embeddings for the KC reviewed at time~$t_n$, $k$ and its neighboring KCs~$i$:  
    $$
    (\bm{h}^k_n)^\prime= \begin{cases}{\left[\bm{h}^k_n, u^{x_n, y_n}\right]} & (i=k) \\ {\left[\bm{h}^k_n, u^i \right]} & (i \neq k \,\, \text{with} \,\, a_{ik} = 1)\end{cases}
    $$
    After aggregating the information from the neighboring KCs, \acro{GKT} updates the hidden states based on the aggregated features and the graph structure:
    $$
    \bm{h}^k_{n+1}= \begin{cases}{ f_\theta (\bm{h}^{k,\prime}_n, \bm{h}^k_n ) } & (i=k) \\ {f_\theta ((\bm{h}^k_n)^\prime, (\bm{h}^i_n)^\prime, \bm{h}^k_n ) } & (i \neq k \,\, \text{with} \,\, a_{ik} = 1).\end{cases}
    $$
    Finally, \acro{GKT} uses an MLP layer to predict the probability of a correct answer at the next time step, $\hat{y}_{n+1}^k = f_\theta(\bm{h}^k_{n+1})$ where $k=x_{n+1}$. 
    
    \item \acro{QIKT} \citep{kt-qikt} focuses on assignments together with KCs, where multiple different assignments can test one KC. Inspired by item-response theory (IRT) \citep{kt-irt}, \acro{QIKT} defines three modules, each parameterized by a neural network, to infer interpretable features, namely assignment-specific knowledge acquisition $\alpha_n$, assignment-specific problem-solving ability $\zeta_{n+1}$, and assignment-agnostic but KC-specific knowledge mastery $\beta_n$. 
    Apart from neural network parameters, \acro{QIKT} learns three sets of assignment-specific embeddings $\{v^{q,0}, v^{q,1}, v^q \}$, which have the same purpose as the KC embeddings defined in \acro{AKT}, namely for correct interactions, incorrect interactions, and assignment identity. Furthermore, another set of KC-specific embeddings~$u^k$ is learned for KC-specific features. 
    \acro{QIKT} uses LSTM and sum pooling to learn the three features based on each learner's history, $\alpha_n:= f_\theta(V_{1:n}, U_{1:n}), \beta_n:= f_\theta (U_{1:n})$, where $V_{1:n}$ and $U_{1:n}$ denote respectively the all the assignments and KC embeddings in the learning history. The problem-solving ability $\zeta_{n+1} := f_\theta(V_{1:n+1}, U_{1:n+1})$ is learned by including the assignment and KC information in the coming interaction. 
    To predict performance, all three features are aggregated and input into the sigmoid function, ~$\hat{y}_{n+1}=\sigma(\alpha_n+\beta_n+\zeta_{n+1})$.
    
\end{enumerate}

\subsubsection{Datasets} \label{appsec:datasets}
Here we describe the datasets that we have used for evaluation (Assist12, Assist17, and Junyi15), articulate the reasons for their selection, and discuss some of the limitations derived from this choice.
\paragraph{Description of the selected datasets}
Assist12 and Assist17 are two subsets of the ASSISTments dataset released by Worcester Polytechnic Institute \citep{assistments}. 
ASSISTments is an online educational tool widely used in U.S. mathematics classes for learners from grades 4 to 12. Predominantly, its users are middle school students (grades 6-8) from Massachusetts or its vicinity. This platform is used for both classroom and homework assignments, and can be used with or without accompanying paper materials. One of the key features of ASSISTments is the immediate feedback provided to students after they answer a question, allowing them to promptly know whether their response was correct. 

Junyi Academy is a non-profit Chinese online education platform. Their Junyi15 data release reports the interactions of more than 72,000 students solving mathematics assignments over a year, totaling 16 million attempts. These interaction logs are provided along with two commissioned annotation sets (`expert' and `crowd-sourced') concerning the structure of 837 KCs in the curriculum. Expert annotations, provided by three teachers, consist of 553 identified prerequisite relations. Crowd-sourced annotations, provided by 51 graduates from senior high school or higher, consist of both prerequisite and similarity evaluations for 1954 KC pairs, with each relation strength rated on a scale from 0 to 9 by at least 3 workers.

\begin{figure}[h]
\centering
    {\includegraphics[height=8in]{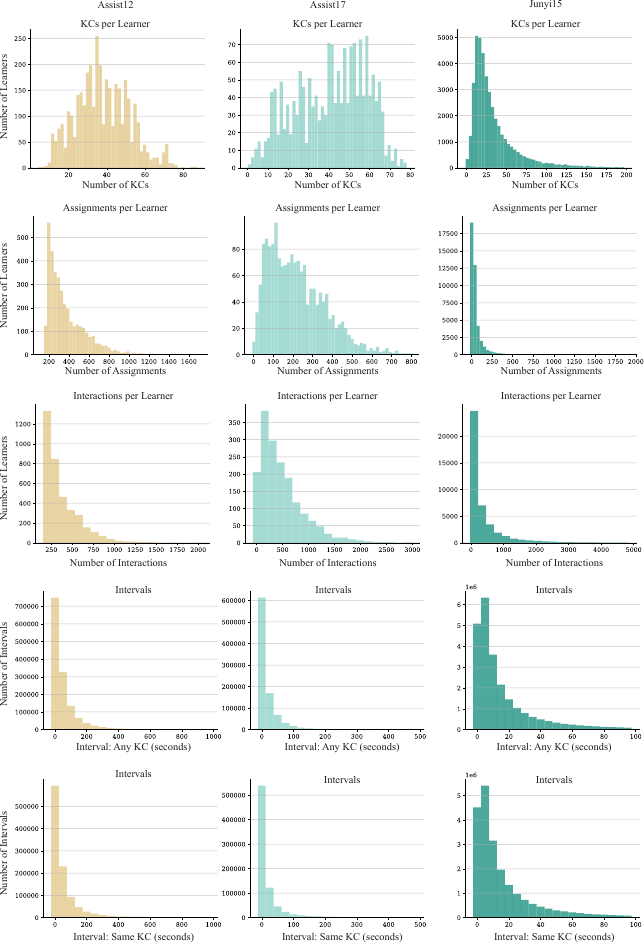}} 
    \caption{Histograms of key features in three datasets, including the number of interactions, KCs, and assignments per learner, and the intervals between two interactions with any KC and with the same KC. 
    }
    \label{appfig:dataset}
\end{figure}

We report the numbers of learners, KCs, assignments, and interactions for each dataset in Table \ref{apptab:datasets}. In Figure \ref{appfig:dataset} we complement this basic characterization with histograms of the number of per-learner interactions, KCs, and assignments, as well as histograms of elapsed time between learner interactions with arbitrary KCs as well as between interactions with the same KC. 
\begin{table}[]
\centering
\scriptsize
\begin{tabular}{l|cc|cc|cc}
\toprule
 & \multicolumn{2}{c|}{Assist12} & \multicolumn{2}{c|}{Assist17} & \multicolumn{2}{c}{Junyi15} \\
 & All & $>$ 50 & All & $>$ 50 & All & $>$ 50 \\
 \midrule
\# Interactions & 6,123,270 & 2,431,788 & 942,816 & 942,489 & 25,925,992 & 23,907,121 \\
\# Learners & 46,674 & 12,443 & 1,709 & 1,697 & 247,606 & 77,655 \\
\# Assignments & 179,999 & 51,866 & 3,162 & 3,162 & 5,174 & 6,174 \\
\# KCs & 265 & 263 & 102 & 102 & 722 & 721 \\
KC Examples & \multicolumn{2}{c|}{\begin{tabular}[c]{@{}c@{}}Rounding\\ Unit Rate\\ Perimeter of a Polygon\end{tabular}} & \multicolumn{2}{c|}{\begin{tabular}[c]{@{}c@{}}substitution\\ fraction-division\\ prime-number\end{tabular}} & \multicolumn{2}{c}{\begin{tabular}[c]{@{}c@{}}matrix\_basic\_distance\\ circles\_and\_arcs\\ arithmetic\_means\end{tabular}} \\ \bottomrule
\end{tabular}
\caption{Overview of Educational Datasets Assist12, Assist17, and Junyi15, including the number of interactions, learners, assignments, and KCs from overall log data (All) and the log data including learners with more than 50 interactions ($>$ 50). }
\label{apptab:datasets}
\end{table}

\paragraph{Criteria for dataset selection}
In order to empirically test our model of learning in structured domains, we sought datasets from domains with a clear prerequisite structure that provide (1) identifiable KC labels, and (2) interaction times with sufficient temporal resolution. 
In domains where prerequisite relations between KCs are strong, the correct learning order is key for performance, so that performance data be used to uncover structural relations. Additionally, the dependencies in these domains can be identified independently by human annotators, which we use to validate model inferences about the knowledge structure.

\begin{enumerate}

 \item Identifiable KC labels. Some datasets do not identify the specific KC reviewed at an interaction, but rather a more general assignment or task that could involve multiple unspecified KCs. While this assignment structure can be explicitly modeled (e.g. our baselines \acro{AKT}, \acro{HKT}, and \acro{QIKT}), and we do intend to extend our model in future work to cover this setting, here we intentionally avoided modeling assignment features and concentrated directly on the underlying KCs and their dependencies, which requires KC identities.
 
 \item Timestamped interactions with high temporal resolution. A resolution in the order of seconds or less is essential to adequately track the initial phases of the forgetting process, and to model structural influences that depend on the precise order of KC presentation (see Eq.~\ref{eq:ou-process-mean}). %

\end{enumerate}

Following these criteria, we had to exclude the Statics2011 dataset due to a lack of identified KCs. 
The Assistments2009 and Assistments2015 datasets lack timestamps entirely, while the 15-minute temporal resolution of the Junyi20 dataset is too coarse for our purposes. This leaves us with Assist12, Assist17, and Junyi15 as appropriate choices to evaluate KT on structured domains. Besides abundant interaction data, Junyi15 provides human-annotated KC relations that, while noisy, offer an invaluable reference to compare the inferred prerequisite graphs.

\paragraph{Limitations} The selection of datasets is limited by design to structured domains, where we can more appropriately put to the test our structure-aware model. We acknowledge that when KCs are largely unrelated (e.g., general knowledge trivia) the inference of prerequisite structure may confer no real advantage. Mathematics, in contrast, provides an ideal testing ground, but more interaction datasets from other domains (e.g., biology, chemistry, linguistics...) and learning stages (primary school, college) are needed for a more representative assessment of the role of structure in learning. In the future, we intend to extend our model to accommodate a broader range of datasets, addressing, in particular, the common case where a single interaction, such as an assignment or a task, is associated with multiple KCs, which entails a more complex interplay of KCs than is displayed in our current dataset selection \citep{wang2020diagnostic}.

%% file: tables/models.tex
\begin{table}[h] 
    \scriptsize
    \centering
    \caption{Models. \textit{\# Emb/KC} is the number of learnable embeddings per KC. 
    \textit{Forgetting} is the functional form of memory decay, with exponential ($\exp$) decay the most common.}
    \vspace{-5pt}
    \label{tab:models}
    \small
    \begin{tabular}{lcccccccc}
    \toprule
        Feature & \acro{HLR}/\acro{PPE}  & \acro{DKT} & \acro{DKTF} & \acro{HKT}  & \acro{AKT} & \acro{GKT} & \acro{QIKT}  & \acro{PSI-KT} \\ \midrule
        \# Emb/KC& \xmark  & 2 & 2 & 6 & 6 & 3 & 1 & 1 \\
        Forgetting & $\exp$ & \xmark & $\exp$ & Hawkes & \xmark & \xmark & \xmark & OU \\
    \bottomrule
    \end{tabular}
\end{table}
\vspace*{-\parskip}

%% file: appendix/7_3_groupkt_model.tex
\subsection{\acro{PSI-KT} model architecture} \label{appsec:groupkt-architecture}
\subsubsection{Network details}
\begin{figure}[h]
\centering
    {\includegraphics[height=2.8in, trim={0 0 2cm 0},clip]{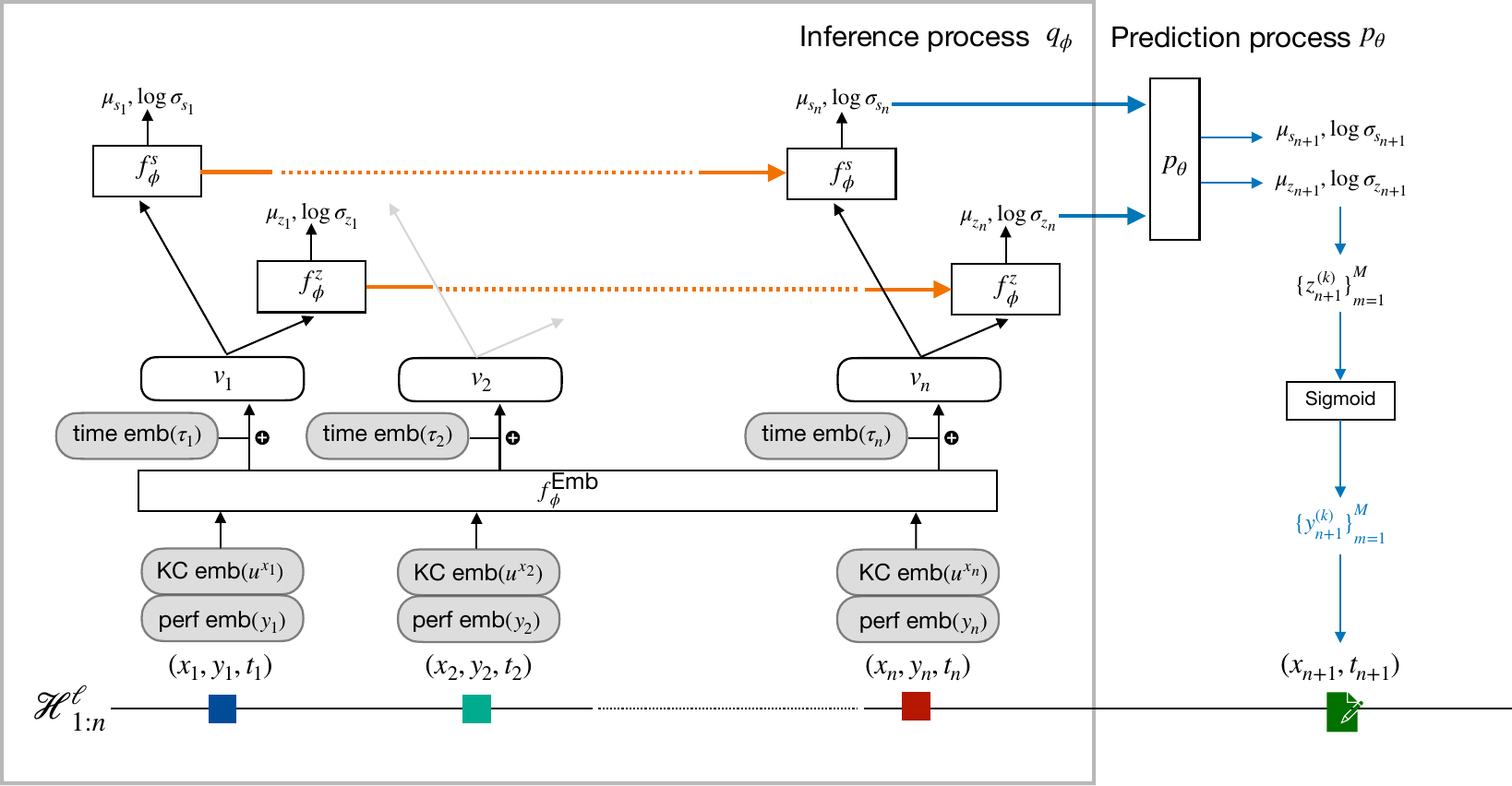}} 
    \caption{
    Inference model of \acro{PSI-KT} with an example of a single learner's history as the input. Note that all parameters~$\phi, \theta$ are shared across learners. 
    Grey backgrounds mark inputs. Right rectangles are neural networks' layers and rounded rectangles designate features. 
    Layer~$f_{\phi}^\text{Emb}$ maps the input features (time~$\tau$, KC~$u$, and performance $y$, described in the text below) into embedding vectors. Layers~$f_{\phi}^s$ and~$f_{\phi}^z$ output the parameters of the variational posterior distribution. These are all part of the inference networks and parameterized by~$\phi$ (surrounded by the grey box).
    The orange arrow is only applicable for inference on entire learning histories. 
    Blue arrows represent the prediction stage, where during prediction~$M$ samples are drawn from the predicted distribution based on~$\mu_{z_{n+1}}$ and~$\log \sigma_{z_{n+1}}$.
    }
    \label{appfig:model_detail}
\end{figure}

In this section, we introduce the detailed architecture of our \acro{PSI-KT} model and its hyperparameters.
The inference network consists of an embedding network~$f_{\phi}^\text{Emb}$, the cognitive traits encoder~$f_{\phi}^s$, and the knowledge states encoder~$f_{\phi}^z$. The weights of these interconnected networks collectively constitute the inference parameters~$\phi$. 

\paragraph{Interaction embedding network. } The network~$f^{\text{Emb}}_\phi$ extracts features from the learning history tuples~$\mathcal{H}_{1:N}^\ell = \{x_n, y_n, t_n \}^\ell_{1:N}$, combining information about interaction time, KC identity and performance.

The \emph{KC identity embedding} for KC~$x_n$ corresponds to the learned embedding~$u^{x_n}$, which is part of the generative model that parameterizes the graph structure. 
The \emph{performance embedding} is obtained by expanding the scalar value~$y_n$ into a vector~$\Vec{y}_n$ with the same dimensionality as the time and KC embeddings so that the performance features will be represented on an equal footing. We then concatenate the KC embedding~$u^{x_n}$ with the performance embedding~$\Vec{y}_n$.
The \emph{interval embedding} is a positional encoding \citep{transformer},~${\rm PE}_n = (\sin \alpha(\tau_n); \cos \alpha(\tau_n))$. This embedding approach accommodates intervals spanning different timescales, from minutes to weeks. 

Thus, the joint embedding for a learning interaction is given by~$v_n = f_{\phi, \text{Emb}}([u^{x_n}; \Vec{y}_n]) + {\rm PE}_n$, inspired by the transformer architecture \citep{transformer}. 

\paragraph{Latent state encoder. } The network~$f^z_\phi$ infers the parameters of the variational posterior distribution~$q_\phi(\bm{z}_{1:n})$. Since learning histories do not have a pre-determined length, we use an LSTM \citep{lstm} as the inference architecture. At each time point, we extract the hidden states in the LSTM, $h_{\bm{z}_{1:n}} = \text{LSTM}(v_{1:n})$. 
Meanwhile, in the continual learning setting, information about the history is already encoded and available in the variational parameters for the last time step~$\phi_{n-1}$, so we use a multi-layer perceptron (MLP), $h_{\bm{z}_n} = \text{MLP}(v_n)$. 
Finally, another MLP \citep[similar to the encoder in][]{kingma2014autoencoding} takes the hidden states~$h_{\bm{z}_n}$ at every time point as inputs and produces the mean $\mu_{\bm{z}_n} \in \mathbb{R}^K$ and log-variance~$\log \sigma_{\bm{z}_n} \in \mathbb{R}^K$ for knowledge states~$\bm{z}_n$.

\paragraph{Latent trait encoder. }The network~$f^s_\phi$ infers the parameters of the variational posterior distribution~$q_\phi(s_{1:n})$. The resulting approximate posterior distribution enables the sampling of learner-specific traits to facilitate personalized predictions. 
One immediately obvious approach is to use the same architecture of~$f^z_{\phi}$. However, the unimodal Gaussian prior over the latent variables cannot account for the diversity of cognitive trait combinations that we expect to find across learners in diverse cohorts. What we need is to allow for multimodality in the distribution of $s$ over all learners. 

There is work on factorizing the joint variational posterior as a combination of isotropic posteriors, using a mixture of $M$~experts \citep[MoE;][]{shi2019variational}, i.e., $q_{\phi}(s_{1:n}^\ell \,|\, \mathcal{H}_{1:n}^\ell)=\nicefrac{1}{M} \sum_m q_{\phi_m}(s_{1:n}^\ell \,|\, \mathcal{H}_{1:n}^\ell)$, assuming the different modalities are of comparable complexity. However, this may lead to over-parameterization.
Instead, inspired by \citet{dilokthanakul2016deep}, we opt for a mixture of Gaussians as a prior distribution that generalizes the unimodal Gaussian prior and provides multimodality. By assuming that the observed data arises from a mixture of Gaussians, determining the category of a data point becomes equivalent to identifying the mode of the latent distribution from which the data point originates. This approach allows us to partition our latent space into distinct categories. With these discrete variables, it is no longer possible to directly apply the reparameterization trick. To solve this inference challenge, we modify the standard VAE architecture by incorporating the Gumbel-Softmax trick \citep{jang2016categorical}.
We employ an LSTM network, taking history embeddings~$v_{1:n}$ as inputs and generating one of $C$ category labels through the Gumbel-Softmax technique, denoted as~$w = \text{LSTM}(v_{1:n})\in {\rm Cat}(\bm{\pi})$. Here $\operatorname{Cat}(\boldsymbol{\pi})$ represents the categorical distribution with probabilities ~$\boldsymbol{\pi} \in \Delta^C$.  
Simultaneously, we capture hidden states at each time point as~$h_{z_{1:n}}$. Subsequently, we utilize an MLP to process both the category label and hidden states as input, producing the mean~$\mu_{s_n} \in \mathbb{R}^4$ and log-variance~$\log \sigma_{s_n} \in \mathbb{R}^4$ of latent states~$s_n$ for each time point. 

Table~\ref{apptab:model-hyperparams} presents an overview of the \acro{PSI-KT} model architecture and hyperparameters used for all experiments.

\input{tables/app-model-hyperparams}

%% file: tables/app-model-hyperparams.tex
\begin{table}[]
    \centering
    \caption{\acro{PSI-KT} architecture and hyperparameters. FC$(a,b)$ represents a fully connected layer with input dimension~$a$ and output dimension~$b$; $K$ represents the number of KCs, different across datasets; $C$ represents the number of categories in the mixture of Gaussians for $s$ (we use $C=10$ in our experiments); the semicolon $;$ separates connected layers, while the slash $/$ separates the layer architecture for inference on entire histories from the continual learning set-up, where different.
    }
    \label{apptab:model-hyperparams}
    \scriptsize
    \begin{tabular}{lccc}
        \toprule
        & Inputs \& Dim & Hidden Layers & Outputs \\ \midrule
        $f_{\phi}^\text{Emb}$ 
            & \begin{tabular}[c]{@{}c@{}}KC Emb \& 16\\ Perf Emb \& 16\end{tabular} 
            & \begin{tabular}[c]{@{}c@{}}FC (32, 16)\\ LeakyReLU(0.2)\\ FC (16, 16)\end{tabular} 
            & $v_n$ \\ \midrule
        $f_{\phi}^z$ 
            & $v_n$ \& 16 
            & \begin{tabular}[c]{@{}c@{}}LSTM (16, 32) / FC (16, 32) \\ FC (32, 16); LeakyReLU(0.2) \\FC (16, 16); LeakyReLU(0.2) \\ FC (16, $K$); FC (16, $K$) \end{tabular} & $\mu_{\bm{z}_n}, \log \sigma_{\bm{z}_n}$ \\ \midrule
        $f_{\phi}^s$ 
            & $v_n$ \& 16 
            & \begin{tabular}[c]{@{}c@{}} FC (16, 32)\\ FC (32, 16); LeakyReLU(0.2)\\ FC(16, 64); LeakyReLU(0.2)\\ GumbelSoftmax(FC(64, $C$))\\ FC(32 + $C$, 64); LeakyReLU(0.2) \\ FC(64, 16); LeakyReLU(0.2) \\ FC(16);  FC(64, 4) \end{tabular} & $\mu_{s_n}, \log \sigma_{s_n}$ \\
        \bottomrule
    \end{tabular}
\end{table}

%% file: appendix/7_4_exp_pred.tex
\subsection{Prediction and generalization experiments details} \label{appsec:exp-prediction}

\subsubsection{Within-learner prediction results and training hyperparameters} \label{appsec:exp-prediction-detail}
In our prediction experiments, we employ a supervised training approach. For each learner, the first 10 interactions from their learning history are used for training, with the subsequent 10 interactions used as the test set. To report results, we reserve 20\% of the learners as a validation set.
We employ the Adam optimizer \citep{adam} with an initial learning rate of 0.005 and apply gradient clipping with a threshold of 10.0. We use a linear decay schedule for the learning rate, halving it every 200 epochs. Additionally, we maintain a consistent batch size of 32 across models.

In Figure~\ref{fig:performance-results} in the main text, we present the average accuracy curves for comparison. For a more comprehensive overview of our training protocols, including accuracy, F1-score, and their standard deviation across 5 random seeds, please refer to the detailed results provided in Appendix Tables~\ref{apptab:prediction-results-accuracy} and \ref{apptab:prediction-results-f1}. 

In our baseline models, the original approach was to predict a single time point in the future using all available historical data. However, we believe that relying solely on short-term predictions is insufficient for capturing long-term trends in learners' performance, which is crucial for making accurate recommendations for customized learning materials. Moreover, it's often impractical to assume that we can always access ground-truth data for immediate predictions.
Therefore, we predict 10 time points into the future, using the predicted performances as inputs for each step. In other words, instead of using ground-truth data, if the model can predict~$\hat{y}_n$ based on all previous training data~$y_{n^\prime<n}$, we incorporate the predicted performance along with the historical data~$[y_{n^\prime<n};\hat{y}_n]$ to predict~$\hat{y}_{n+1}$.
\input{tables/app-prediction-results-rebuttal}

In the evaluations, we chose to focus on prediction and generalization on a small group of learners, with numbers ranging from 100 to 1,000. This decision is based on the reality that, in educational settings, large datasets are not always available or practical. Additionally, little data is key in practical ITS to minimize the number of learners on an experimental treatment, to mitigate the cold-start problem, and extend the usefulness of the model to classroom-size groups. To provide ITS with a basis for adaptive guidance and long-term learner assessment, we always predict the 10 next interactions.

In order to ensure a fair evaluation of deep learning models and to avoid biasing our results, we expanded our dataset to include over 1,000 learners. This expansion was done post-filtering, where we excluded learners with fewer than 50 interactions. Additionally, 20\% of these learners were designated as a validation set. The average accuracy, along with the number of learners and the number of parameters used in each model, is detailed in Table~\ref{apptab:prediction-results-more-learners}.

It's crucial to recognize that deep learning models, despite benefiting from extensive datasets, face specific challenges. Firstly, \acro{PSI-KT} has remarkable predictive performance when trained on small cohorts whereas baselines require training data from at least 60k learners to reach similar performance. Secondly, the deployment of these deep learning models in real-time applications is challenging due to their substantial number of parameters.

\input{tables/app-prediction-results-more-learners-rebuttal}

\subsubsection{Between-learner fine-tuning hyperparameters} 
For between-learner generalization, we employ pre-trained models from within learner prediction, where the details can be found in Appendix~\ref{appsec:exp-prediction-detail}. These models are trained using data from 100 learners, and we retain the one that achieved the highest prediction accuracy on the validation set. Then, predictions are made by randomly selecting 100 learners from the group that were not included in the training or validation sets. 

In the experiment without fine-tuning, we directly apply the pre-trained models to unseen out-of-sample learners and present the results in Table~\ref{tab:generalization-results}. This entails using the pre-trained models to predict the next 10 interactions for out-of-sample learners based on their first 10 interactions as input.

In the fine-tuning experiment, we perform fine-tuning for each model using a batch size of 32. Additionally, we also set aside 20\% of the learners as a validation set during this process to save the model that achieves the highest accuracy after fine-tuning.
For baseline models, \acro{HLR}, \acro{PPE}, and \acro{HKT}, which are comprised entirely of learner-independent and KC-dependent parameters, we conduct fine-tuning for all of these parameters. In this scenario, we use the pre-trained models as the initial weight values for the fine-tuning process.
Conversely, for models \acro{DKT}, \acro{DKTF}, and \acro{AKT}, we perform fine-tuning specifically on their KC embedding parameters and the last fully connected layer within the neural network, while keeping the remaining layers frozen during the fine-tuning process.

\subsubsection{Continual-learning results} \label{appsec:continual-learning}
\input{tables/app-continual-results}

In this experiment, we randomly select 100 learners using five different random seeds, and collect their first 100 interactions. Initially, the models are trained using only the first 10 interactions, following the same setup as in the within-learner prediction experiment. Following the initial training, we continuously integrate new interaction data into the training process, introducing one interaction at a time for each learner. 
The model iteratively predicts the subsequent 10 performances. This simulates a common real-world scenario, where learners continually interact with existing or even new KCs. 

In the objective function~$\rm ELBO$ shown in Eq.~\ref{eq:elbo-vcl}, all historical information up to time~$t_n$ has already been fully encoded into the variational parameters~$\phi_n$. Additionally, to allow for the possibility of learners encountering a new KC during their learning journey, we allow for optimization over the KC parameters in the generative model.
As a result, when new interaction data~$\mathcal{I} = (x_{n+1}, y_{n+1}, t_{n+1})^{1:L}$ becomes available at time~$t_{n+1}$, we use the new data to update both the inference model parameters~$\phi_{n+1}$ and the generative model parameters~$U$ and~$M$ which are related to KCs. 

For the baseline models, which are designed to predict performances based on fixed learning histories, there is no need to update the model parameters for each new data point from each learner individually. Instead, when new interaction data, denoted as~$\mathcal{I} = (x_{n+1}, y_{n+1}, t_{n+1})^{1:L}$, becomes accessible at time $t_{n+1}$, we update all the model parameters using all the interaction data collected up to that point, referred to as $\mathcal{H}_{1:n+1}$. This update is performed through 10 gradient descent processes. It is important to note that we do not include an additional validation set to determine when to stop training each model separately. Instead, we aim for a fair comparison among all models, ensuring that they are trained on equal footing with the same limited data and resources available to them.

%% file: tables/app-prediction-results-rebuttal.tex
\begin{table}[t]
    \vspace{-10pt}
    \caption{Accuracies for within-learner prediction across numbers of learners (mean $\pm$ \acro{SEM} across random seeds). 
    }
    \label{apptab:prediction-results-accuracy}
    \centering
    \scriptsize
    \begin{tabular}{llccccccccc}
        \toprule
        Dataset & \# Learners & \acro{HLR} & \acro{PPE} & \acro{DKT} & \acro{DKTF} & \acro{HKT} & \acro{AKT} & \acro{GKT} & \acro{QIKT} & \acro{PSI-KT} \\ \midrule
        & 100
            & .54$_{.03}$ & .65$_{.01}$ & .65$_{.03}$ & .60$_{.01}$ & .55$_{.01}$ & \underline{.67}$_{.02}$ & .63$_{.03}$ & .63$_{.03}$ & \textbf{.68}$_{.02}$ \\
        & 200
            & .55$_{.02}$ & .63$_{.03}$ & .66$_{.02}$ & .62$_{.01}$ & .58$_{.01}$ & \underline{.67}$_{.02}$ & .61$_{.02}$ & .66$_{.02}$ & \textbf{.70}$_{.02}$ \\
        & 300
            & .55$_{.01}$ & .66$_{.01}$ & .67$_{.01}$ & .62$_{.00}$ & .58$_{.01}$ & \underline{.69}$_{.02}$ & .65$_{.02}$ & .65$_{.02}$ & \textbf{.71}$_{.01}$  \\ 
        & 400
            & .55$_{.01}$ & .65$_{.01}$ & \underline{.68}$_{.01}$ & .63$_{.01}$ & .60$_{.02}$ & .67$_{.03}$ & .63$_{.02}$ & .66$_{.01}$ & \textbf{.71}$_{.01}$  \\ 
        & 500
            & .55$_{.01}$ & .64$_{.01}$ & \underline{.67}$_{.01}$ & .63$_{.01}$ & .59$_{.03}$ & \underline{.67}$_{.02}$ & .63$_{.02}$ & .65$_{.02}$ & \textbf{.70}$_{.01}$  \\ 
        \multirow{-6}{*}{{Assist12}} & 1,000
            & .54$_{.00}$ & .65$_{.00}$ & .68$_{.01}$ & .63$_{.01}$ & .60$_{.01}$ & \textbf{.70}$_{.02}$ & .64$_{.01}$ & .64$_{.01}$ & \textbf{.70}$_{.01}$  \\ 
        \midrule
            
        & 100
            & .45$_{.01}$ & .53$_{.02}$ & .57$_{.02}$ & .53$_{.03}$ & .52$_{.03}$ & .56$_{.02}$ & .56$_{.04}$ & \underline{.58}$_{.02}$ & \textbf{.63}$_{.02}$ \\
        & 200
            & .45$_{.01}$ & .53$_{.02}$ & .57$_{.02}$ & .54$_{.02}$ & .54$_{.01}$ & .55$_{.01}$ & .56$_{.02}$ & \underline{.60}$_{.02}$ & \textbf{.63}$_{.01}$ \\
        & 300
            & .46$_{.01}$ & .53$_{.01}$ & .57$_{.02}$ & .55$_{.02}$ & .55$_{.02}$ & .56$_{.04}$ & .58$_{.02}$ & \underline{.61}$_{.01}$ & \textbf{.63}$_{.01}$ \\
        & 400
            & .45$_{.01}$ & .53$_{.01}$ & .56$_{.01}$ & .57$_{.02}$ & .56$_{.02}$ & .56$_{.02}$ & .58$_{.02}$ & \underline{.61}$_{.01}$ & \textbf{.64}$_{.00}$ \\
        & 500
            & .46$_{.01}$ & .53$_{.00}$ & .60$_{.01}$ & .58$_{.01}$ & .54$_{.01}$ & .56$_{.02}$ & .58$_{.01}$ & \underline{.61}$_{.02}$ & \textbf{.63}$_{.01}$ \\
        \multirow{-6}{*}{{Assist17}} & 1,000
            & .44$_{.01}$ & .55$_{.}01$ & .60$_{.01}$ & .57$_{.01}$ & .57$_{.01}$ & .61$_{.01}$ & .60$_{.01}$ & \underline{.63}$_{.01}$ & \textbf{.64}$_{.00}$ \\
        \midrule

        & 100
            & .55$_{.02}$ & .66$_{.03}$ & .79$_{.03}$ & .78$_{.01}$ & .63$_{.02}$ & \underline{.81}$_{.02}$ & .78$_{.02}$ & \underline{.81}$_{.02}$ & \textbf{.83}$_{.02}$ \\
        & 200
            & .57$_{.01}$ & .65$_{.03}$ & .79$_{.01}$ & .78$_{.02}$ & .68$_{.03}$ & \underline{.80}$_{.01}$ & \underline{.80}$_{.01}$ & \underline{.80}$_{.01}$ & \textbf{.84}$_{.01}$ \\
        & 300
            & .56$_{.02}$ & .65$_{.03}$ & \underline{.81}$_{.01}$ & .79$_{.01}$ & .70$_{.01}$ & \underline{.81}$_{.01}$ & .78$_{.02}$ & \underline{.81}$_{.01}$ & \textbf{.85}$_{.01}$ \\
        & 400
            & .61$_{.02}$ & .65$_{.02}$ & .81$_{.01}$ & .80$_{.02}$ & .69$_{.02}$ & \underline{.82}$_{.02}$ & .75$_{.02}$ & .80$_{.01}$ & \textbf{.85}$_{.01}$ \\
        & 500
            & .61$_{.01}$ & .67$_{.02}$ & \underline{.82}$_{.01}$ & .80$_{.02}$ & .70$_{.01}$ & \underline{.82}$_{.02}$ & .78$_{.02}$ & .81$_{.01}$ & \textbf{.85}$_{.01}$ \\
        \multirow{-6}{*}{{Junyi15}} & 1,000
            & .59$_{.01}$ & .66$_{.02}$ & .81$_{.01}$ & .81$_{.00}$ & .69$_{.01}$ & .82$_{.01}$ & .79$_{.02}$ & \underline{.83}$_{.01}$ & \textbf{.85}$_{.01}$ \\
        \bottomrule
    \end{tabular}
    \vspace{-0pt}
\end{table}

\begin{table}[t]
    \vspace{-0pt}
    \caption{F1 scores for within learner prediction across learner numbers (mean $\pm$ \acro{SEM} across random seeds.)
    }
    \label{apptab:prediction-results-f1}
    \centering
    \scriptsize
    \begin{tabular}{llccccccccc}
        \toprule
        Dataset & \# Learners & \acro{HLR} & \acro{PPE} & \acro{DKT} & \acro{DKTF} & \acro{HKT} & \acro{AKT} & \acro{GKT} & \acro{QIKT} & \acro{PSI-KT} \\ \midrule
        & 100
            & .59$_{.02}$ & .77$_{.01}$ & .77$_{.03}$ & .72$_{.01}$ & .64$_{.01}$ & \underline{.79}$_{.02}$ & .76$_{.01}$ & .73$_{.03}$ & \textbf{.80}$_{.01}$ \\
        & 200
            & .60$_{.02}$ & .74$_{.03}$ & \underline{.78}$_{.02}$ & .73$_{.01}$ & .68$_{.01}$ & .76$_{.02}$ & .74$_{.02}$ & .77$_{.02}$ & \textbf{.82}$_{.01}$ \\
        & 300
            & .59$_{.02}$ & .77$_{.01}$ & \underline{.79}$_{.01}$ & .74$_{.00}$ & .69$_{.01}$ & .73$_{.03}$ & .76$_{.02}$ & .77$_{.01}$ & \textbf{.83}$_{.01}$ \\
        & 400
            & .60$_{.02}$ & .77$_{.01}$ & \underline{.79}$_{.01}$ & .74$_{.01}$ & .70$_{.03}$ & .73$_{.03}$ & .75$_{.01}$ & .76$_{.01}$ & \textbf{.83}$_{.01}$ \\
        & 500
            & .60$_{.01}$ & .76$_{.01}$ & \underline{.79}$_{.01}$ & .74$_{.01}$ & .64$_{.10}$ & .74$_{.02}$ & .75$_{.02}$ & .76$_{.01}$ & \textbf{.82}$_{.01}$ \\
        \multirow{-6}{*}{{Assist12}} & 1,000
            & .60$_{.01}$ & .76$_{.00}$ & \underline{.79}$_{.00}$ & .74$_{.01}$ & .71$_{.01}$ & .73$_{.02}$ & .76$_{.01}$ & .76$_{.01}$ & \textbf{.82}$_{.00}$ \\
        \midrule
            
        & 100
            & \underline{.45}$_{.01}$ & .44$_{.01}$ & .42$_{.02}$ & .40$_{.03}$ & .42$_{.03}$ & .40$_{.02}$ & .40$_{.02}$ & .41$_{.02}$ & \textbf{.48}$_{.03}$ \\
        & 200
            & \underline{.45}$_{.01}$ & .44$_{.01}$ & .40$_{.03}$ & .42$_{.01}$ & .43$_{.01}$ & .44$_{.01}$ & .41$_{.02}$ & .43$_{.02}$ & \textbf{.47}$_{.04}$ \\
        & 300
            & \underline{.45}$_{.01}$ & \underline{.45}$_{.02}$ & .40$_{.02}$ & .41$_{.01}$ & .42$_{.03}$ & \underline{.45}$_{.03}$ & .42$_{.01}$ & .44$_{.03}$ & \textbf{.46}$_{.03}$ \\
        & 400
            & .44$_{.01}$ & .44$_{.02}$ & .41$_{.01}$ & .42$_{.02}$ & .43$_{.02}$ & \underline{.45}$_{.03}$ & .42$_{.02}$ & \underline{.45}$_{.01}$ & \textbf{.47}$_{.03}$ \\
        & 500
            & \underline{.46}$_{.01}$ & .45$_{.01}$ & .40$_{.01}$ & .42$_{.00}$ & .40$_{.10}$ & .45$_{.02}$ & .43$_{.02}$ & .45$_{.01}$ & \textbf{.47}$_{.02}$ \\
        \multirow{-6}{*}{{Assist17}} & 1,000
            & .44$_{.01}$ & .44$_{.02}$ & .40$_{.01}$ & .43$_{.00}$ & .43$_{.03}$ & \underline{.46}$_{.02}$ & .43$_{.02}$ & \textbf{.47}$_{.01}$ & \textbf{.47}$_{.04}$ \\
        \midrule

        & 100
            & .53$_{.02}$ & .70$_{.03}$ & .88$_{.02}$ & .87$_{.01}$ & .75$_{.03}$ & \underline{.89}$_{.01}$ & .87$_{.01}$ & \underline{.89}$_{.01}$ & \textbf{.92}$_{.01}$ \\
        & 200
            & .54$_{.02}$ & .71$_{.02}$ & .88$_{.01}$ & .87$_{.01}$ & .80$_{.02}$ & .88$_{.01}$ & .88$_{.01}$ & \underline{.89}$_{.01}$ & \textbf{.91}$_{.01}$ \\
        & 300
            & .53$_{.02}$ & .71$_{.02}$ & .89$_{.01}$ & .88$_{.01}$ & .80$_{.01}$ & \underline{.90}$_{.01}$ & .87$_{.02}$ & .89$_{.01}$ & \textbf{.92}$_{.01}$ \\
        & 400
            & .52$_{.02}$ & .72$_{.03}$ & .89$_{.01}$ & .88$_{.01}$ & .80$_{.01}$ & \underline{.90}$_{.01}$ & .87$_{.02}$ & .89$_{.01}$ & \textbf{.92}$_{.02}$ \\
        & 500
            & .53$_{.01}$ & .70$_{.02}$ & \underline{.89}$_{.01}$ & .88$_{.01}$ & .74$_{.08}$ & .88$_{.01}$ & .86$_{.01}$ & \underline{.89}$_{.01}$ & \textbf{.92}$_{.01}$ \\
        \multirow{-6}{*}{{Junyi15}} & 1,000
            & .52$_{.01}$ & .71$_{.02}$ & \underline{.90}$_{.01}$ & .89$_{.00}$ & .80$_{.02}$ & \underline{.90}$_{.01}$ & .85$_{.01}$ & \underline{.90}$_{.01}$ & \textbf{.93}$_{.00}$ \\
        \bottomrule
    \end{tabular}
    \vspace{-5pt}
\end{table}

%% file: tables/app-prediction-results-more-learners-rebuttal.tex
\begin{table}[t]
    \vspace{-10pt}
    \caption{Accuracy score in within-learner prediction with all learners in each dataset (mean $\pm$ \acro{SEM} across random seeds). 
    }
    \label{apptab:prediction-results-more-learners}
    \centering
    \scriptsize
    \begin{tabular}{llccccccccc}
        \toprule
        Dataset & \# Learners & \acro{HLR} & \acro{PPE} & \acro{DKT} & \acro{DKTF} & \acro{HKT} & \acro{AKT} & \acro{GKT} & \acro{QIKT} & \acro{PSI-KT} \\ \midrule
        Assist17 & 1,358  & .46$_{.00}$ & .55$_{.00}$ & .58$_{.01}$ & .55$_{.01}$ & .57$_{.01}$ & .60$_{.01}$ & .60$_{.01}$ & \underline{.61}$_{.01}$ & \textbf{.64}$_{.00}$ \\
         Assist12 & 9,954 & .44$_{.02}$ & .47$_{.00}$ & \underline{.69}$_{.00}$ & .66$_{.00}$ & .66$_{.00}$ & 68$_{.01}$ & \textbf{.70}$_{.00}$ & .68$_{.00}$ & \textbf{.70}$_{.01}$ \\
            
         Junyi15 & 62,124 & .65$_{.01}$ & .71$_{.01}$ & .85$_{.00}$ & .85$_{.00}$ & .84$_{.01}$ & \textbf{.86}$_{.01}$ & \textbf{.86}$_{.02}$ & \textbf{.86}$_{.00}$ & .85$_{.01}$  \\
        \bottomrule
    \end{tabular}
    \vspace{-0pt}
\end{table}

%% file: tables/app-continual-results.tex
\begin{table}[h]
    \vspace{-10pt}
    \caption{Continual learning accuracy. We report accuracy in predicting 10 subsequent outcomes. \# Data indicate the number of interactions from each learner for training. }
    \label{apptab:continual-results-accuracy}
    \centering
    \scriptsize
    \begin{tabular}{llccccccccc}
        \toprule
        Dataset & \# Data & 10 & 20 & 30 & 40 & 50 & 60 & 70 & 80 & 90 \\ \midrule
        & \acro{HLR}
            & .54$_{.03}$ & .57$_{.08}$ & .58$_{.08}$ & .59$_{.09}$ & .57$_{.10}$ & .56$_{.07}$ & .54$_{.07}$ & .55$_{.06}$ & .57$_{.08}$ \\
        & \acro{PPE}
            & .65$_{.01}$ & .55$_{.07}$ & .53$_{.07}$ & .52$_{.08}$ & .54$_{.06}$ & .57$_{.06}$ & .59$_{.06}$ & .61$_{.06}$ & \underline{.69}$_{.04}$ \\
        & \acro{DKT}
            & .65$_{.03}$ & .66$_{.07}$ & .64$_{.06}$ & \underline{.68}$_{.05}$ & \underline{.69}$_{.04}$ & \underline{.66}$_{.05}$ & \underline{.66}$_{.05}$ & \underline{.68}$_{.03}$ & .65$_{.01}$ \\
        & \acro{DKTF}
            & .60$_{.01}$ & \underline{.67}$_{.04}$ & \underline{.65}$_{.04}$ & .64$_{.04}$ & .66$_{.03}$ & .62$_{.06}$ & .61$_{.04}$ & .63$_{.02}$ & .63$_{.02}$ \\
        & \acro{HKT}
            & .55$_{.01}$ & .56$_{.05}$ & .62$_{.04}$ & .62$_{.05}$ & .63$_{.02}$ & .60$_{.02}$ & .61$_{.03}$ & .61$_{.02}$ & .62$_{.02}$ \\
        & \acro{AKT}
            & .67$_{.02}$ & .66$_{.04}$ & .62$_{.04}$ & .61$_{.04}$ & .61$_{.05}$ & .65$_{.02}$ & .62$_{.02}$ & .61$_{.02}$ & .63$_{.02}$ \\
        & \acro{GKT}
            & .65$_{.02}$ & .62$_{.02}$ & .62$_{.01}$ & .64$_{.05}$ & .65$_{.04}$ & .65$_{.03}$ & \underline{.66}$_{.06}$ & .65$_{.05}$ & .65$_{.05}$ \\
        & \acro{QIKT}
            & \textbf{.70}$_{.02}$ & .63$_{.01}$ & .64$_{.02}$ & .63$_{.01}$ & .62$_{.03}$ & .62$_{.01}$ & .62$_{.02}$ & .62$_{.02}$ & .63$_{.01}$ \\
        \multirow{-9}{*}{{Assist12}} & \acro{PSI-KT}
            & \underline{.68}$_{.02}$ & \textbf{.70}$_{.03}$ & \textbf{.68}$_{.03}$ & \textbf{.72}$_{.03}$ & \textbf{.75}$_{.02}$ & \textbf{.73}$_{.03}$ & \textbf{.74}$_{.02}$ & \textbf{.74}$_{.02}$ & \textbf{.74}$_{.02}$ \\
        \midrule
        & \acro{HLR}
            & .45$_{.01}$ & .46$_{.07}$ & .45$_{.07}$ & .53$_{.06}$ & .55$_{.08}$ & .57$_{.06}$ & .55$_{.06}$ & .55$_{.04}$ & .54$_{.03}$ \\
        & \acro{PPE}
            & .53$_{.02}$ & .52$_{.06}$ & .52$_{.06}$ & .52$_{.07}$ & .52$_{.06}$ & .52$_{.05}$ & .51$_{.05}$ & .54$_{.04}$ & .56$_{.04}$ \\
        & \acro{DKT}
            & .57$_{.02}$ & .52$_{.05}$ & .52$_{.05}$ & .52$_{.06}$ & .59$_{.04}$ & .57$_{.05}$ & .60$_{.04}$ & \textbf{.63}$_{.02}$ & .59$_{.03}$ \\
        & \acro{DKTF}
            & .53$_{.03}$ & .58$_{.05}$ & .54$_{.05}$ & \underline{.58}$_{.05}$ & .58$_{.04}$ & .55$_{.05}$ & .56$_{.05}$ & .56$_{.04}$ & \underline{.61}$_{.02}$ \\
        & \acro{HKT}
            & .52$_{.03}$ & .57$_{.04}$ & \underline{.60}$_{.03}$ & \textbf{.60}$_{.03}$ & \textbf{.62}$_{.02}$ & \underline{.61}$_{.03}$ & \underline{.61}$_{.02}$ & .60$_{.02}$ & \underline{.61}$_{.02}$ \\
        & \acro{AKT}
            & .56$_{.02}$ & .53$_{.05}$ & .52$_{.06}$ & .54$_{.04}$ & .53$_{.04}$ & .53$_{.02}$ & .50$_{.03}$ & .51$_{.03}$ & .57$_{.02}$ \\
        & \acro{GKT}
            & \underline{.63}$_{.02}$ & \underline{.59}$_{.05}$ & .54$_{.04}$ & \textbf{.60}$_{.04}$ & .56$_{.03}$ & .54$_{.02}$ & .57$_{.02}$ & .58$_{.02}$ & .58$_{.03}$ \\
        & \acro{QIKT}
            & \textbf{.65}$_{.02}$ & .58$_{.03}$ & .59$_{.03}$ & .56$_{.05}$ & .58$_{.03}$ & .56$_{.02}$ & .58$_{.02}$ & .58$_{.01}$ & .56$_{.02}$ \\
        \multirow{-9}{*}{{Assist17}} & \acro{PSI-KT}
            & \underline{.63}$_{.02}$ & \textbf{.62}$_{.04}$ & \textbf{.65}$_{.04}$ & \textbf{.60}$_{.05}$ & \underline{.60}$_{.05}$ & \textbf{.62}$_{.05}$ & \textbf{.62}$_{.04}$ & \underline{.62}$_{.04}$ & \textbf{.64}$_{.03}$ \\
        \midrule
        & \acro{HLR}
            & .55$_{.02}$ & .43$_{.06}$ & .42$_{.06}$ & .44$_{.05}$ & .60$_{.04}$ & .63$_{.04}$ & .63$_{.03}$ & .63$_{.04}$ & .64$_{.03}$ \\
        & \acro{PPE}
            & .66$_{.03}$ & .67$_{.06}$ & .64$_{.06}$ & .64$_{.05}$ & .62$_{.04}$ & .63$_{.05}$ & .60$_{.05}$ & .60$_{.03}$ & .61$_{.03}$ \\
        & \acro{DKT}
            & .79$_{.03}$ & \underline{.80}$_{.04}$ & .78$_{.04}$ & .76$_{.05}$ & .77$_{.04}$ & .75$_{.04}$ & \textbf{.84}$_{.04}$ & .73$_{.02}$ & .74$_{.01}$ \\
        & \acro{DKTF}
            & .78$_{.01}$ & .74$_{.05}$ & .77$_{.05}$ & .74$_{.06}$ & .71$_{.05}$ & .71$_{.04}$ & .74$_{.03}$ & .71$_{.03}$ & .72$_{.02}$ \\
        & \acro{HKT}
            & .63$_{.02}$ & .63$_{.08}$ & .69$_{.07}$ & .67$_{.07}$ & .70$_{.04}$ & .73$_{.04}$ & .73$_{.03}$ & \underline{.79}$_{.02}$ & \underline{.84}$_{.02}$ \\
        & \acro{AKT}
            & .81$_{.02}$ & .79$_{.04}$ & .78$_{.05}$ & \underline{.79}$_{.04}$ & .75$_{.04}$ & .75$_{.03}$ & .76$_{.03}$ & .74$_{.02}$ & .74$_{.03}$ \\
        & \acro{GKT}
            & .82$_{.01}$ & \underline{.80}$_{.02}$ & .78$_{.03}$ & .78$_{.03}$ & \textbf{.79}$_{.04}$ & \underline{.79}$_{.03}$ & .79$_{.03}$ & \underline{.79}$_{.02}$ & .80$_{.02}$ \\
        & \acro{QIKT}
            & \textbf{.84}$_{.00}$ & \underline{.80}$_{.02}$ & \underline{.80}$_{.05}$ & \underline{.78}$_{.03}$ & .78$_{.04}$ & \textbf{.81}$_{.03}$ & .80$_{.02}$ & .78$_{.01}$ & \textbf{.85}$_{.02}$ \\
        \multirow{-9}{*}{{Junyi15}} & \acro{PSI-KT}
            & \underline{.83}$_{.02}$ & \textbf{.81}$_{.04}$ & \textbf{.81}$_{.04}$ & \textbf{.80}$_{.04}$ & .77$_{.06}$ & \textbf{.81}$_{.04}$ & \underline{.82}$_{.04}$ & \textbf{.83}$_{.03}$ & \underline{.84}$_{.03}$ \\
        \midrule
    \end{tabular}
    \vspace{-10pt}
\end{table}

%% file: appendix/7_5_exp_learner.tex
\subsection{Learner-specific representations analysis} \label{appsec:interpretability-learner}
In this experiment, we examine temporal latent features (learner representations) derived from baseline models. When considering baseline models, it is noteworthy that only \acro{DKT}, \acro{DKTF}, \acro{AKT}, and \acro{QIKT} incorporate learner-specific temporal embedding vectors. While \acro{HKT} utilizes temporal embeddings, all these embeddings originate from global parameters associated with KCs, rendering them non-learner-specific. 
Consequently, our comparative analysis focuses exclusively on \acro{PSI-KT} compared with \acro{DKT}, \acro{DKTF}, \acro{AKT}, and \acro{QIKT}.

We initially present comprehensive results in Table \ref{apptab:interpretability-learner}, complementing Table \ref{tab:interpretability-learner} from Section \ref{sec:exp-interpretability}, wherein only the results from the best-performing baseline models are displayed. Subsequent sections will detail the experimental setups and metrics employed.

\input{tables/app-interpretability-learner}

\subsubsection{Experimental setup for specificity} \label{appsec:interpretability-learner-identifiability}
In personalized learning, we assume each learner has a unique cognitive profile shaped by past experiences and educational contexts. Our first step is to connect learner representations $s_n^\ell$ with these inherent learner-specific cognitive traits, i.e., the \emph{specificity} of learners given corresponding representations.

To quantify specificity, we employ mutual information, denoted as $\textrm{MI}(s;\ell):= \text{H}(s) - \text{H}(s|\ell)$ among all learners, as a measure of the information shared between learner identities and learner representations. The detailed computation of the metric~$\textrm{MI}(s;\ell)$ is outlined as follows: 
\begin{align} \label{appeq:interpretability-mi}
   \textrm{MI}(s;\ell) & = \text{H}(s) - \text{H}(s|\ell) \nonumber \\
   & = -\int p(s) \log p(s) - \frac{1}{L} \sum_\ell \int p(s|\ell) \log p(s|\ell) \nonumber \\
   & = \frac{1}{2} \bigl( D (1+\log2\pi) + \log|\Sigma_s| \bigr) 
        - \frac{1}{2L} \sum_\ell \bigl( D(1+\log2\pi) + \log|\Sigma_{s^\ell}| \bigr) \nonumber \\
   & = \frac{1}{2} \bigl( \log |\Sigma_s| - \frac{1}{L} \sum_\ell \log |\Sigma_{s^\ell}| \bigr).
\end{align}
Here~$\Sigma_s$ and~$\Sigma_{s^\ell}$ are the covariance matrices obtained from fitting learner representations from all $L$~learners or, respectively, a single learner with a Gaussian distribution, and $D$ is the dimensionality of learner representations. 
In experiments, we begin by randomly selecting 1,000 learners from each dataset and then extracting their first 50 interactions for training. To determine when to stop training effectively, we set aside a validation set of 20\% of learners, which amounts to 200 learners in our case. This setup mirrors our approach in the prediction experiment.
The metric~$\textrm{MI}(s;\ell)$ is calculated for the learners in the training set. Since our goal here is to evaluate the model's capacity to distill representations $s^\ell$ that uniquely identify learners, there is no need for a test set. Note that the baseline models have higher-dimensional learner representations (16 dimensions in our experiments), potentially allowing them to capture more information.

\subsubsection{Experimental setup for consistency}
We proceed with a supplementary \emph{consistency} analysis to determine, among the shared information quantified in specificity, whether the learner representations capture intricate learner attributes or merely reflect transient dynamic fluctuations. 
In the experiment, we split the interaction data of each learner into five separate groups, i.e., subsets. Each subset contains 30 interactions. These specific sizes were chosen to ensure we have both enough learners for robust training and enough interactions in subsets to estimate covariance matrices for our metrics. We thus exclude learners who have engaged in fewer than 150 interactions.

To form subsets, we find out the average presentation time of each KC and assign the KCs to separate subsets, so that the overall average interaction time is as similar as possible across subsets. With this, we aim to wash out, to the extent possible with the limited amount of data, systematic biases in the partition induced by the dependence of learner representations on time. 

The mutual information metric~$\mathbb{E}_{\ell_\textrm{sub}}  \textrm{MI} (s^{\ell};\ell_\textrm{sub}) := \mathbb{E}_{\ell_\textrm{sub}} \left[ \text{H} (s| \ell)- \text{H} (s| \ell_\textrm{sub}) \right]$, employed in the consistency experiments, undergoes the a similar derivation process to Eq.~\ref{appeq:interpretability-mi}.
\begin{align}
    \mathbb{E}_{\ell_\textrm{sub}}  \textrm{MI} (s^{\ell};\ell_\textrm{sub}) 
    & = \frac{1}{L}\sum_\ell \bigl(\text{H}(s|\ell) - \frac{1}{5}\sum_{\ell_{\text{sub}}} \text{H}(s|\ell_{\text{sub}})\bigr) \nonumber \\
    & = \frac{1}{L}\sum_\ell \Bigl( - \mathbb{E} \bigl[ \log \mathcal{N} \left( \mu_{s^\ell}, \Sigma_{s^\ell}^2 \right) \bigr] 
    + \frac{1}{5} \sum_{\ell_{\text{sub}}} \mathbb{E} \bigl[ \log \mathcal{N} \left( \mu_{s^{\ell_{\text{sub}}}}, \Sigma_{s^{\ell_{\text{sub}}}}^2 \right) \bigr] \Bigr) \nonumber \\
    & = \frac{1}{L}\sum_\ell \bigl( \log |\Sigma_{s^\ell}| - \frac{1}{5} \sum_{\ell_{\text{sub}}} \log |\Sigma_{s^{\ell_{\text{sub}}}}| \bigr).
\end{align}

We fit each sub-learner separately and quantify the divergence metric~$\mathbb{E}_{\ell_\textrm{sub}}  \textrm{MI} (s^{\ell};\ell_\textrm{sub})$ between learners and their sub-learners.
A lower value of~$\mathbb{E}_{\ell_\textrm{sub}} \textrm{MI} (s^{\ell};\ell_\textrm{sub})$ suggests a higher degree of consistency, reflecting the difficulty in distinguishing between sub-learners and their corresponding overarching learners given learner representations. 
Overall, Table~\ref{tab:interpretability-learner} shows that the learner representations of \acro{PSI-KT} provide comparable learner specificity and superior consistency. The lower consistency displayed by baseline models suggests that most of the representational capacity available in their higher-dimensional representations might be spent on capturing learner-unspecific characteristics of the training sample.

\subsubsection{Experimental setup for disentanglement}
With the insights gained from specificity, our analysis progresses to evaluating to what extent learner-specific representations, are disentangled. 
Disentanglement in machine learning has been characterized as the process of isolating and identifying distinct, independent, and informative generative factors of variation in the data \citep{bengio2013representation}. 

In our disentanglement experiments, we use the same setup for specificity, and compute the discrepancy~$ D_\textrm{KL} (s \| \ell)$ based on 50 interactions of 1,000 learners. Our approach bears similarity to \citep{kim2018disentangling}, but we relax the unrealistic assumption of independent representations. In real-world scenarios, independence in cognitive attributes is not a priority. To assess how much information about learner identity is present in the covariance across the representation dimension, we use the divergence between full trait-vector entropy and diagonal learner-conditional trait-vector entropy. 

The discrepancy~$ D_\textrm{KL} (s \| \ell)$ is estimated by the full entropy of representations~$\text{H}(s)$ and the diagonal elements of the covariance matrix in the conditional entropy~$\text{H}(s|\ell)$
\begin{align}
    D_\textrm{KL} (s \| \ell) := \text{H}(s)_{\text{full}} - \text{H}(s|\ell)_{\text{diag}} = \frac{1}{2} \left(\log|\Sigma_s| - \frac{1}{L} \sum_\ell \sum_{i=1}^D \log {(\Sigma_{s|\ell})}_{ii} \right).
\end{align}
Small non-diagonal elements of the covariance matrix in~$\text{H}(s)$ suggest low cross-correlations. This can be interpreted as a form of disentanglement. As illustrated in the third row of Table~\ref{tab:interpretability-learner}, the representations from \acro{PSI-KT} consistently exhibit a higher degree of disentanglement across all datasets. 

\subsubsection{Mixed-effect linear regressions in operational interpretability} \label{appsec:interpretability-learner-mixed-effect}
Mixed-effects regression extends linear regression to handle data with hierarchical or clustered structures, such as repeated measurements from the same subjects (learners in our case). 
Taking one of our experiments as an example, we conduct regressions based on~$y_n^\ell \sim \Tilde{\mu}^{\ell,k}_n +(1 \mid \text{learner})$. Here, $y_n^\ell$ represents the dependent variable and $\Tilde{\mu}^{\ell,k}_n$ is a predictor variable at time~$t_n$. Also, $(1 \mid \text{learner})$ represents the random intercept associated with each learner. This random intercept accounts for variability between learners that cannot be explained by the fixed effect $\Tilde{\mu}^{\ell,k}_n$. In other words, it accounts for the fact that different learners might have different biases in their responses, allowing us to capture a more robust estimate of the group-level effect. 

For regression calculations, we use the models trained in the prediction experiments, as described in Section~\ref{appsec:exp-prediction}. For consistent comparisons with specificity experiments, we opt for models trained on a group of 1,000 learners. This experiment goes beyond a simple sanity check (as in Sec.~\ref{appsec:interpretability-learner-identifiability}), so we use the testing data. This choice aligns with our objective of using operational interpretability to gain insights and inform future controlled experiments with unseen data. We use pre-trained models, specifically \acro{DKT}, \acro{DKTF}, \acro{AKT}, and our \acro{PSI-KT} model, selected based on their accuracy scores on the validation data.

To fairly compare with baseline models, we investigate whether any dimensions within the learner representations capture behaviors similar to our interpretable cognitive traits. Thus, we perform regression for each dimension within the learner representations of the baseline models. 
While Figure~\ref{fig:interpretability-learner-regression} in the main paper presents the regression results concerning the dimension featuring the most pronounced correlation among baseline models, we provide a complete list of dimensions that exhibit significant relationships with the behavioral data in Table~\ref{apptab:interpretability-learner-correlation}. 

\input{tables/app-interpretability-learner-regression}

\paragraph{Performance decay and forgetting rate}
\begin{figure}[t] 
    \centering
    {\includegraphics[height=5.5in]{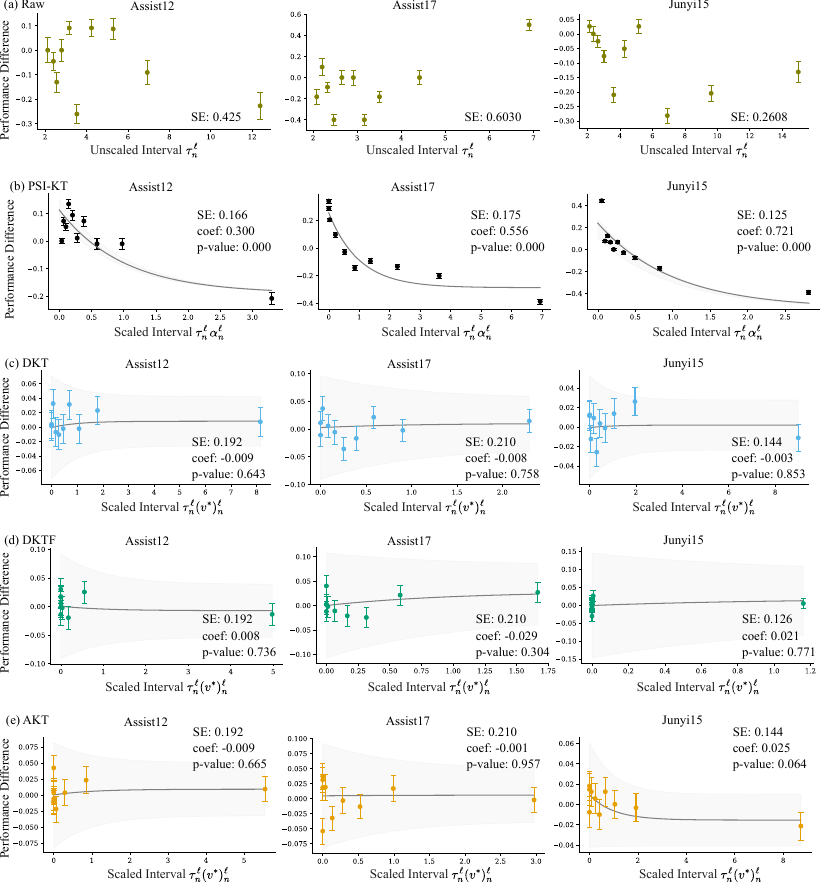}}
    \caption{The mixed-effect regressions of performance decay~$\Delta y_n^\ell$ vs. scaled interval~$\tau_n^\ell \alpha_n^\ell$ (scaled interval with the best dimension in baselines $\tau_n^\ell (v^*)_n^\ell$). The first row (a) shows the unscaled interval in the raw data. The aggregate standard error over 10 bins (SE), the regression coefficient (coef), and its $p$-value are reported in each panel. }
    \label{appfig:alpha-regression}
\end{figure}
To analyze the exponential decay of learner performances over time, we first show the relationship between performance decay~$\Delta y_n^\ell$ and the raw time difference~$\tau_n^\ell$, which is divided into 10 bins. We select bin centers to ensure an equal number of data points in each bin. This binning approach helps minimize the impact of outliers and ensures a balanced representation of data within each bin. 
Also, we show the relationship between decay~$\Delta y_n^\ell$ and the time difference scaled by the corresponding forgetting rate~$\alpha_n^\ell$ at each time point, or each dimension of learner representations in the baseline models. 
We assume that if the forgetting rate~$\alpha_n^\ell$ is meaningful for each time interval and effectively controls the decay, then the standard error of behavior data~$\Delta y_n^\ell$ within each bin should be smaller than the error of binning raw time differences. This indicates that the decay is better described as a function of $\alpha^\ell_n \tau^\ell_n$ than as a function of $\tau^\ell_n$ alone. 
We also compute the standard error for each dimension of learner representations in the baseline models, and we show the dimension~$v^*_n$ with the lowest standard error in Figure~\ref{appfig:alpha-regression}. 
Then, we perform mixed-effect regression over the exponential term~$\exp(-\alpha_n^\ell\,\tau_n^\ell)$ (or $\exp(-v^*_n\,\tau_n^\ell)$ in baselines) to assess how well learner representations predict performance decay (as an exponential function). The results show that at least one dimension in the learner representations groups certain behavioral data and reduces the standard errors. However, none of these dimensions exhibit a statistically significant relationship with the behavioral data. 

\paragraph{Initial performance and long-term mean}
\begin{figure}[t] 
    \centering
    {\includegraphics[height=4.4in]{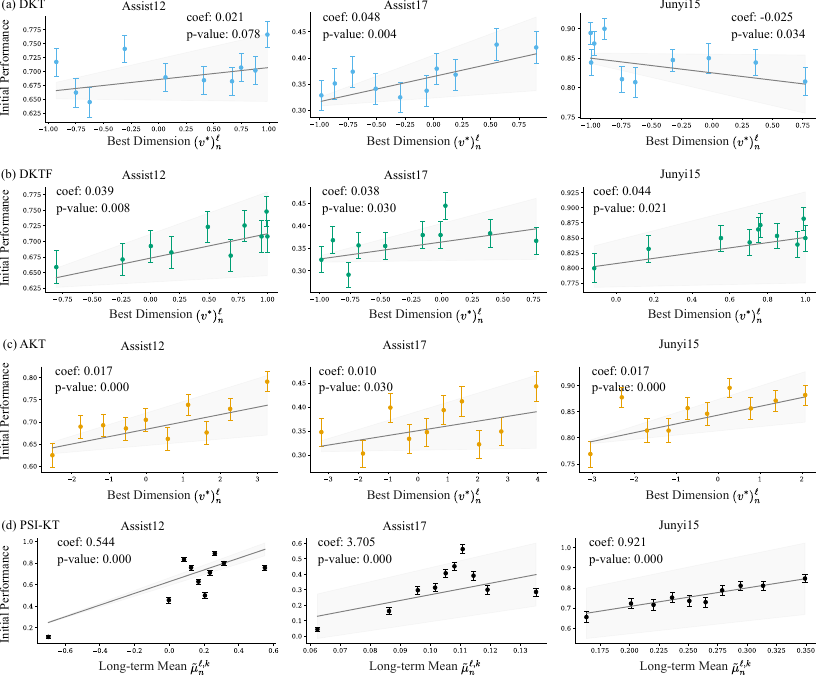}}
    \caption{The mixed-effect regressions of initial performance~$y_n^\ell$ vs. long-term mean~$\Tilde{\mu}_n^{\ell,k}$ (the best dimension~$ (v^*)_n^\ell$ in baselines). The aggregate standard error over 10 bins (SE), the regression coefficient (coef) and its $p$-value are reported in each panel. }
    \label{appfig:mu-regression}
\end{figure}
We conducted a mixed-effect regression analysis between the initial performance and the long-term mean, with the results presented in Figure~\ref{appfig:mu-regression}. These results indicate that, except \acro{DKT} on the Assist12 dataset, at least one dimension in the baseline learner representations predicts initial performance.
It is important to note that none of the dimensions exhibit a stronger effect compared to the identified trait~$\Tilde{\mu}_n^{\ell,k}$ in our \acro{PSI-KT}. Additionally, we note that embedding dimensions in baselines are trained in a permutation-invariant manner, suggesting that these models can't route any particular generative factor of variation in the data (e.g. a behavioral signature) to a specific dimension.

\paragraph{Prerequisite transfer ability and learning variances}
In our experiments, we sought to correlate two additional cognitive traits – transfer ability~$\gamma$ and learning volatility~$\sigma$ - with behavioral data. This task proved more complex than assessing the forgetting rate and long-term mean because assessing transfer ability requires reliable annotations of prerequisite relations and learning volatility can be connected to many unconstrained factors during the learning process. 

Regarding transfer ability, our hypothesis posits that given the identified prerequisite KC~$i$ for KC~$j$, a higher transfer ability~$\gamma^\ell_n$ suggests an increased likelihood of correctly transitioning from one KC~$i$ to KC~$j$. We calculate this transition probability~$p(j^+ \,|\, i^+)^\ell_n$ by observing the frequency of correct responses to KC~$i$ followed by correct responses to KC~$j$ up to a certain time~$t_n$. This implies that learners with greater transfer abilities are more likely to answer questions related to KC~$i$ correctly after mastering KC~$i$. However, this approach depends on accurately identifying prerequisite relationships between KCs. Therefore, we utilized the Junyi15 dataset, which includes expert-annotated and crowd-sourced prerequisite graphs, for our regression analysis.
For learning volatility~$\sigma^\ell_n$, we connect the average squared mean~$(\bar{\sigma}^\ell)^2$ for each learner with the variance in their performance~$\text{Var}(y^\ell_{1:n})$. 

\begin{figure}[t] 
    \centering
    {\includegraphics[height=2.2in]{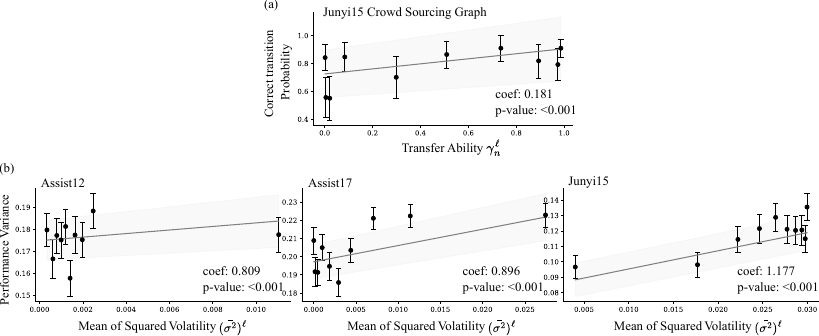}}
    \caption{The mixed-effect regressions of transfer ability~$\gamma$ with behavioral correct transition probability (a), and learning volatility~$\sigma$ with the variance in learning performances (b). We report the regression coefficient (coef) and its $p$-value in each panel, and each point illustrates the mean ($\pm$ SEM) of the corresponding decile. }
    \label{appfig:sigma-gamma}
\end{figure}

In Figure~\ref{appfig:sigma-gamma}, we present the results of our mixed-effect regression analyses. Each regression demonstrates a significant relationship. However, due to the sparsity of the expert-annotated graph, we do not have enough data to fit the regression model effectively. Thus we choose to use the crowd-sourcing graphs and consider the edge existence if the edge weight is above 0.5. 

\subsubsection{Visualization of knowledge states}
In this section, we display the curve of inferred knowledge states within the Junyi15 dataset. We chose sequences where the involved skills are linked by established prerequisite relations. Two such prerequisites were identified: 'alternate interior angles' as a prerequisite for 'corresponding angles', and 'number properties terminology' for 'properties of numbers'. These prerequisites were determined based on crowd-sourcing annotations, where the average score for the annotated relation exceeded half. 
We note that \acro{PSI-KT} can estimate knowledge states at all times and not just interaction times, which allows us to use natural time in the abscissae and display knowledge states with curves instead of using the discrete color maps common in the KT literature.

\begin{figure}[t] 
    \centering
    {\includegraphics[height=1.4in]{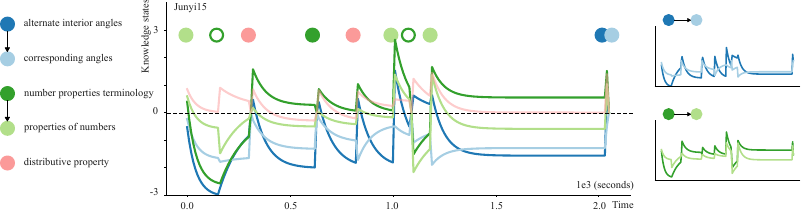}}
    \caption{An example of inferred sequential knowledge states in the Junyi15 dataset. }
    \label{appfig:knowledge-states-visualization}
\end{figure}

%% file: tables/app-interpretability-learner.tex
\begin{table}[]
    \vspace{-0pt}
    \centering
    \caption{Specificity, consistency, and disentanglement. }
    \label{apptab:interpretability-learner}
    \vspace{-5pt}
    \scriptsize
    \begin{tabular}{llccccc} 
    \toprule
    Metric & Dataset & \acro{DKT} & \acro{DKTF} & \acro{AKT} & \acro{QIKT} & \acro{PSI-KT} \\  \midrule
    \multirow{3}{*}{\begin{tabular}[l]{@{}l@{}} Specificity \\ $\textrm{MI}(s;\ell)$ $\uparrow$ \end{tabular}  } 
        & Assist12 & \textbf{8.83} & 6.62 & 6.45 & 2.47 & \underline{8.40} \\
        & Assist17 & 8.08 & 7.50 & \textbf{10.05} & 2.95 & \underline{9.98} \\
        & Junyi15 & 12.75 & \underline{13.50} & 13.34 & 4.09 & \textbf{14.37} \\ \midrule
    \multirow{3}{*}{\begin{tabular}[l]{@{}l@{}} $\text{Consistency}^{-1}$ \\ $\mathbb{E}_{\ell_\textrm{sub}}  \textrm{MI} (s^{\ell};\ell_\textrm{sub})$ $\downarrow$ \end{tabular}} 
        & Assist12 & 14.13 & 12.24 & 20.15 & \underline{8.35} & \textbf{7.48} \\
        & Assist17 & 14.95 & 13.11 & 24.47 & \underline{6.35} & \textbf{6.35} \\
        & Junyi15 & 13.10 & 17.81 & 22.15 & \underline{7.66} & \textbf{5.00} \\ \midrule
    \multirow{3}{*}{\begin{tabular}[l]{@{}l@{}} Disentanglement \\ $D_\textrm{KL} (s \| \ell)$ $\uparrow$ \end{tabular}} 
         & Assist12 & -1.64 & 0.38 & -8.17 & \underline{2.31} & \textbf{7.42} \\ 
         & Assist17 & -3.01 & -0.44 & -9.81 & \underline{0.56} & \textbf{8.39} \\ 
         & Junyi15 & -0.62 & \underline{4.96} & -6.65 & 1.57 & \textbf{11.49} \\
    \bottomrule
    \end{tabular}
    \vspace{-10pt}
\end{table}

%% file: tables/app-interpretability-learner-regression.tex
\begin{table}[t] 
    \centering
    \caption{(regression coefficient, $p$-value) tuples for performance difference and initial performance across models and latents' dimensions. If there is no significant dimension in one model and dataset ($p> 0.05$), we show the dimension with the highest regression coefficient. Bold values indicate the dimension and the baseline model with the highest statistically significant linear relationship in one dataset, with which we show the regression results in Figures~\ref{appfig:alpha-regression} and~\ref{appfig:mu-regression}.}
    \label{apptab:interpretability-learner-correlation}
    \scriptsize
    \begin{tabular}{llcccc}
    \toprule
    \begin{tabular}[l]{@{}l@{}}Behavioural \\ signature\end{tabular} & Dataset & \acro{DKT} & \acro{DKTF} & \acro{AKT} & \acro{PSI-KT} \\ 
    \midrule
    \multirow{3}{*}{\begin{tabular}[l]{@{}l@{}}Performance \\ difference\end{tabular}} 
        & Assist12 
            & (\underline{-.009}, .643) & (.008, .736) & (\underline{-.009}, .665) & (\textbf{.300}, <.001) \\
        & Assist17 
            & (-.008, .758) & (\underline{-.029}, .304) & (-.001, .957) & (\textbf{.556}, <.001) \\
        & Junyi15 
            & (-.003, .853) & (.021, .771) & (\underline{.025}, .064) & (\textbf{.721}, <.001) \\
    \midrule 
    \multirow{3}{*}{\begin{tabular}[l]{@{}l@{}}Initial \\ performance\end{tabular}} 
        & Assist12 
            & (.021, .078) & (\underline{.039}, .008) & (.017, <.001) & (\textbf{.544}, <.001) \\
        & Assist17 
            &  (\underline{.048}, .004)  &  (.038, .030) & (.010, .030) & (\textbf{3.705}, <.001) \\
        & Junyi15 
            & (-.025, .034) &  (\underline{.044}, .021) & (.017, <.001) & (\textbf{.921}, <.001) \\ 
     \bottomrule
    \end{tabular}
\end{table}

%% file: appendix/7_6_exp_graph.tex
\subsection{Graph inference analysis} \label{appsec:graph}

\subsubsection{Details of the metrics for ground-truth graph comparison} \label{appsec:graph-metric-detail}
Here we report comprehensive evaluations of the alignment of the inferred graphs with the human-annotated graphs in the Junyi15 dataset under different metrics.

As discussed in Section~\ref{sec:exp-interpretability-graph}, the Junyi15 dataset provides two types of graph annotations - crowd-sourced similarity and prerequisite ratings (with 1,954 rated edges), as well as more sparse expert-annotated prerequisite relations (837 edges). We use the following metrics to compare graph representations learned by each model against these annotations:

\begin{enumerate}

     \item \textbf{Mean Reciprocal Rank} (MRR). We compare the inferred graph with the expert-annotated using the MRR, defined as~$|K|^{-1}\textstyle{\sum_{i=1}^{|K|}(\mathrm{rank}(i))^{-1}}$, where $K$~is the total number of KCs. We compute the rank of each expert-identified prerequisite relation~$i \rightarrow k$ in the relevant sorted list of inferred probabilities~$\{a^{jk}\}_{j=1}^K$ and take the harmonic average. 
     
    \item \textbf{Jaccard Similarity} (JS) is a classic measure of similarity between two sets, defined as the size of the intersection of set A and set B (i.e., the number of common elements) over the size of the union of set A and set B (i.e., the number of unique elements): ${\rm JS}(A, B)=\sfrac{|A \cap B|}{|A \cup B|}$. Here we define the edge sets by thresholding weights at half the scale (0.5 for probability-scaled weights, 5 for the average of the 1-9 crowd-sourced rating).
    
    \item \textbf{Negative Log-likelihood} (nLL) of edge weights given crowd-sourced annotations. The crowd-sourced annotations provide multiple 1-9 ratings per node pair. One set of annotations rates the strength of the directed prerequisite relations, whereas the other just rates the undirected similarity of the pair of nodes. We normalize the ratings from 0 to 1 and fit them with a Gaussian distribution. Then we compute the log-likelihood of the inferred edge probability under the Gaussian. The variance of the Gaussian accounts for inter-rater disagreements when comparing a model's inferred edge probability with the mean edge rating.

    \item  \textbf{Linear Regression Coefficient} between edge weights and the causal support (details are in Sec.~\ref{appsec:causal-support}) from node~$i$ to node~$k$ on correctness of~$k$ if having correct interactions on~$i$. We compute the causal support for transitions of every KC pair. However we remove the causal support of pairs of KCs that have only one transition in the dataset to avoid adding noise to our estimate. 
    
\end{enumerate}

\subsubsection{Quantitative comparison results on the Junyi15 dataset} \label{appsec:graph-junyi15-gt-results}
Note that the graphs of baselines are based on KC embeddings (as in Sec.~\ref{appsec:baseline-model}), and thus there is no edge directionality. 
For the baselines that have at least two embeddings for each KC, we can use to compute the directed edges, since one embedding for KC will end up in a symmetric structure adjacency matrix (\acro{DKT}, \acro{DKTF}, \acro{HKT}, \acro{AKT}).
Thus, to conduct a fair comparison with the baseline models, we leniently compute edge weights based on every combination of KC embeddings. For example, in \acro{DKT}, there are two embeddings~$u^{k,0}, u^{k,1} \in \mathbb{R}^D$ representing incorrect interactions and correct interactions on KC~$k$, respectively, and embeddings are shared across all learners. We compute the edge weights~$a^{ik}$ based on two different combinations here, both $a^{ik} := u^{i,0\intercal} u^{k,1}$ and $a^{ik} := u^{i,1\intercal} u^{k,0}$, and report the graph with the best results. When extracting undirected graphs, we concatenate all KC embeddings to compute $a^{ik} := (u^{i,0}+u^{k,1})^\intercal (u^{i,0}+u^{k,1})$, in order to reflect all available information from all KC embeddings. 
In the case of baselines with a single embedding per KC, such as \acro{QIKT}, or those using a parameterized undirected graph, like \acro{GKT}, we allow their inferred graphs to be less accurate. This means that for these models, the presence of an edge between two KCs is deemed correct if there is a directed edge from either direction in annotated graphs, without the necessity for these edges to accurately indicate the directionality.
We then compute the weights by min-max normalization. This normalization is necessary for computing the log-likelihood, where we also use a threshold of 0.5 to determine whether there is an edge when the comparison requires binary edges. 

In Table~\ref{tab:graph-comparison-results}, we show the comparison of ground-truth prerequisite graphs and inferred graphs from our \acro{PSI-KT}, and the best baseline models on the Junyi15 dataset under the different metrics. These results demonstrate that our inferred prerequisite graph outperforms others when compared with crowd-sourced and expert-annotated graphs under different metrics. 

Here we show all of the comparison results, including a comparison of the similarity (undirected) graphs and the prerequisite (directed) graphs on four metrics in Table~\ref{apptab:graph-comparison-results}. We do not report MRR ranking scores for similarity graphs because the ground-truth similarity graph does not contain an expert-annotated version. 

\input{tables/app-graph-comparison-results}

\subsubsection{Causal support}  \label{appsec:causal-support}
Causal induction is the problem of inferring underlying causal structures from data. Here, we use a Bayesian framework \citep{griffiths2009theory, griffiths2005structure} to infer a singular cause-and-effect relationship between all pairs of KCs, asking how performance on one node influences performance on another, and whether the strength of the causal relationship corresponds to our inferred prerequisite graph. In this context, we model the relationship between a candidate cause~$C$ and a candidate effect~$E$ (i.e., a pair of KCs), assuming an ever-present background cause~$B$ (i.e., the learner's general ability and the influence of other nodes). The objectives are to determine the probability of a causal relationship between~$C$ and~$E$, known as \emph{causal support} (Eq.~\ref{appeq:causal-support}).

In our prerequisite graph scenario, we assume that if KC~$i$ is a prerequisite of KC~$k$, the correctness on KC~$i$ contributes to the correctness on KC~$k$. This implies the presence of a prerequisite relationship between KC~$i$ and KC~$k$, signified by a causal link between their correctness levels.
Consequently, for every pair of nodes, a candidate cause~$C$ corresponds to performance~$y_n^i=1$ at time~$t_n$, and an effect~$E$ corresponds to $y_{n+1}^k=1$ at time~$t_{n+1}$, with inputs from all remaining nodes relegated to the background cause~$B$. 

When examining elemental causal induction, we adhere to the following two-step procedure: 
i) We establish the nature of the relationship through causal graphical models, and
ii) we quantify the strength of the relationship, provided it exists, as a problem of inferring structural parameters. 
In the subsequent text, $C$ and $E$ variables are denoted using uppercase letters, while their specific instances are represented using lowercase letters. Specifically, ${\rm c}^+$ and ${\rm e}^+$ indicate the presence of the cause and effect (i.e., correct performance), whereas ${\rm c}^-$ and ${\rm e}^-$ signify their absence (i.e., incorrect performance). 

\begin{figure}[t]
\centering
    {\includegraphics[height=7.7in]
    {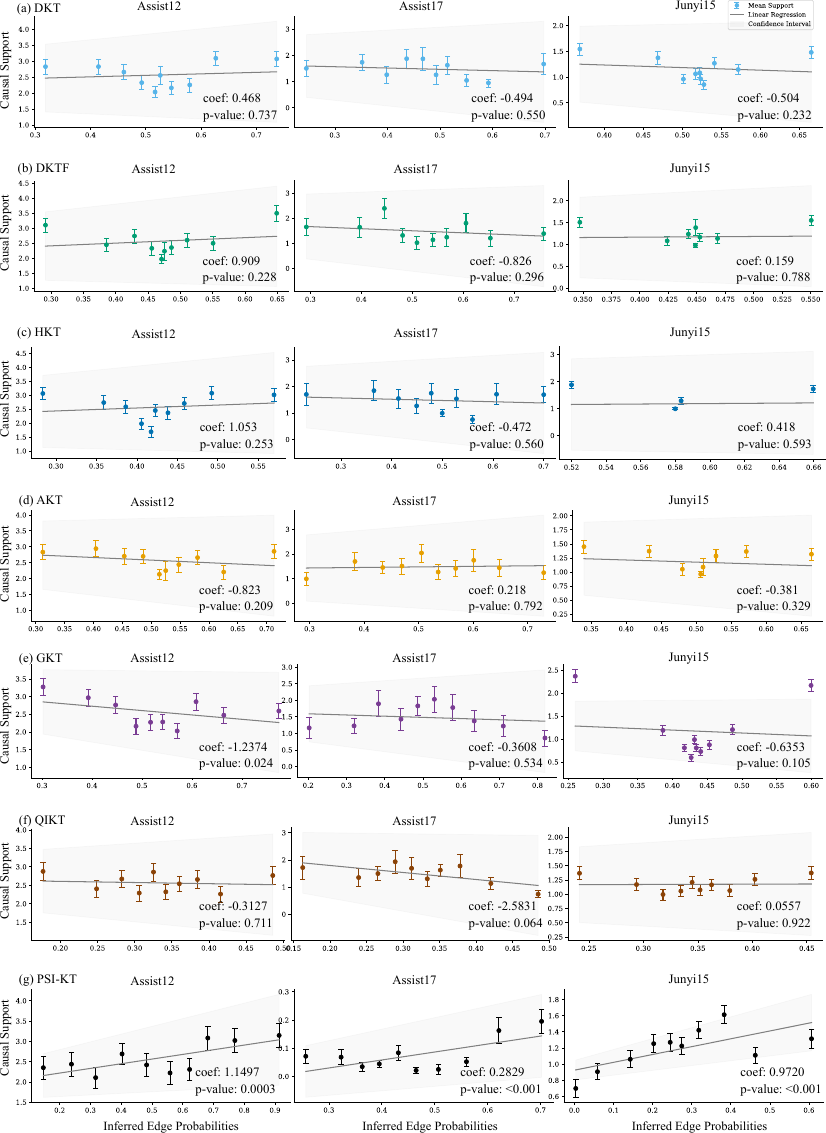}
    } 
    \caption{Linear regressions relating causal support to the inferred edges for baseline models \captiona~\acro{DKT}, \captionb~\acro{DKTF}, \captionc~\acro{HKT}, \captiond~\acro{AKT}, \captione~\acro{GKT}, \captionf~\acro{QIKT}, and \captiong~\acro{PSI-KT}. The $x$-axis represents the normalized edge weights inferred by the respective baselines. The coefficient (coef) and its $p$-value are reported in the lower right of each panel. 
    } 
    \label{appfig:graph-causal-support-regression}
    \vspace{-2em}
\end{figure}

\paragraph{Causal graphical models.}
Causal graphical models are a formalism for learning and reasoning about causal relationships \citep{glymour2019review}. Nodes in the graph represent variables, and directed edges represent causal connections between those variables. 
To identify whether a causal relationship exists between a pair of variables, we consider two directed graphs denoted Graph 0 $G_{C \nrightarrow E}: B \rightarrow E$ and Graph 1 $G_{C \rightarrow E}: B \rightarrow E \leftarrow C$, as shown in Figure~\ref{fig:graph-visualization}b. 
Thus, $G_{C \nrightarrow E}$~represents the null hypothesis that there is no relationship between~$C$ and~$E$ (i.e., the effect~$E$ can be accounted for by background cause~$B$), while~$G_{C \rightarrow E}$ represents the alternative hypothesis that the causal relationship exists. 

In our case, the cause~$C$ and the effect~$E$ are equivalent to KC~$i$ and KC~$k$, respectively, for every pair of KCs. The process of inferring the underlying structure between KC~$i$ and KC~$k$, whether the learners' behavioral learning history~$\mathcal{H}$ are generated by~$G_{i \nrightarrow k}$ or $G_{i \rightarrow k}$, can be cast in a Bayesian framework \citep{griffiths2009theory,griffiths2005structure}. \emph{Causal support} quantifies the degree of evidence present in the data~$\mathcal{H}$ that favors Graph 1~$G_{i \rightarrow k}$ over Graph 0~$G_{i \nrightarrow k}$:
\begin{align} \label{appeq:causal-support}
    \text{support} =\log \frac{P(\mathcal{H} \,|\, G_{C \rightarrow E})}{P(\mathcal{H} \,|\, G_{C \nrightarrow E})} = \log \frac{P(\mathcal{H} \,|\, G_{i \rightarrow k})}{P(\mathcal{H} \,|\, G_{i \nrightarrow k})}.
\end{align}
Intuitively, the joint presence of the cause and effect, i.e., correctness on KC~$i$ followed by correctness on KC~$k$, offers support for a causal link from node~$i$ to node~$k$. Conversely, the absence of the cause, i.e., incorrectness on KC~$i$ but is followed by correctness on KC~$k$, presents evidence against the notion that KC~$i$ is a prerequisite for KC~$k$.

\paragraph{Causal support.}
Causal graphical models depict dependencies using conditional probabilities. Defining these probabilities entails parameterizing each edge, and this parameterization determines the functional expressions that govern causal relationships.

For Graph 0~$G_{i \nrightarrow k}$ and Graph 1~$G_{i \rightarrow k}$, we define~$P_0(y_{n+1}^k=1 \,|\, B) = \omega_0$ and~$P_1(y_{n+1}^k=1 \,|\, y_n^i=1)=\omega_1$ respectively. In other words, the probability of correctness on KC~$k$ given just background causes is $\omega_0$, and the probability of correctness on KC~$k$ given previous correctness on KC~$i$ is $\omega_1$; and when both prerequisite KC~$i$ and background causes are present, they have independent opportunities to produce the effect. 

For \textbf{Graph 0~$G_{C \nrightarrow E}$}, the sole parameter~$\omega_0$ denotes the likelihood of the effect being present given the background cause
\begin{align} \label{appeq:background-cause-omega0}
    P_0({\rm e}^+ \,|\, b^+; \omega_0)=\omega_0.
\end{align}
The corresponding likelihood for the data~$\mathcal{H}$ given Graph 0~$G_{i \nrightarrow k}$ is accomplished by integrating over all possible parameters~$\omega_0$ with a uniform prior over~$\omega_0$:
\begin{align}
    P(\mathcal{H} \,|\, G_{i \nrightarrow k})
    & =\int_0^1 P_0(\mathcal{H} \,|\, \omega_0, G_{i \nrightarrow k})P(\omega_0 \,|\, G_{i \nrightarrow k})\mathrm{d} \omega_0 \nonumber \\
    & = \int_0^1 \omega_0^{N({\rm e}^+)}(1-\omega_0)^{N({\rm e}^-)} \mathrm{d} \omega_0 \nonumber \\
    & = {\rm Beta}(N({\rm e}^+)+1, N({\rm e}^-)+1) \nonumber \\
    & = {\rm Beta}(N(y_{n+1}^k=1)+1, N(y_{n+1}^k=0)+1).
\end{align}
Here ${\rm Beta}()$ is the beta function, and $N({\rm e}^+)$ and $N({\rm e}^-)$ are the marginal frequencies of the effects.

For \textbf{Graph 1 $G_{i \rightarrow k}$}, the likelihood of the effect is given by:
\begin{align} \label{appeq:candidate-cause-omega1}
    P_1({\rm e}^+ \,|\, b, c; \omega_0, \omega_1) = 1 - (1-\omega_0)^b (1-\omega_1)^c,
\end{align}
where $\omega_0$ again defines the influence of the background cause, and the additional parameter $\omega_1$ defines the influence of the cause. Here~$b$ and~$c$ are binary, which means if cause~$C$ exists, then~$c=1$. 
We compute the likelihood of the data~$P(\mathcal{H} \,|\,G_{i \rightarrow k})$ by integrating over parameters~$\omega_0$ and~$\omega_1$. Each parameter value is defined by a prior probability, which when combined with the likelihood of the data, yields a joint posterior distribution over data and parameters for the structure. To determine the observed data likelihood for Graph 1~$G_{i \rightarrow k}$, we have 
\begin{align} \label{appeq:likelihood-graph1-original}
    P(\mathcal{H} \,|\, G_{i \rightarrow k}) 
    & =\int_0^1 \int_0^1 P_1(\mathcal{H} \,|\, \omega_0, \omega_1, G_{i \rightarrow k}) P(\omega_0, \omega_1 \,|\, G_{i \rightarrow k}) \mathrm{d} \omega_0 \mathrm{~d} \omega_1 \nonumber \\
    & =\int_0^1 \int_0^1  \prod_{e, c} P_1(e \,|\, c, {\rm b}^{+} ; \omega_0, \omega_1)^{N(e, c)} P(\omega_0, \omega_1 \,|\, G_{i \rightarrow k}) \mathrm{d} \omega_0 \mathrm{~d} \omega_1.
\end{align}
Here $N(e, c)$~represents the number of occurrences. 
To compute~$\prod_{e, c} P_1(e \,|\, c, {\rm b}^{+} ; \omega_0, \omega_1)^{N(e, c)}$, we iterate over all possible sets of~$(e,c)$. 
Based on Eq.~\ref{appeq:candidate-cause-omega1}, we get:
\begin{align}
    \prod_{e, c} P_1(e \,|\, c, {\rm b}^+ ; \omega_0, \omega_1)^{N(e, c)} 
    = & P_1({\rm e}^+ \,|\, {\rm c}^+, b^+ ; \omega_0, \omega_1)^{N({\rm e}^+, {\rm c}^+)} P_1({\rm e}^+ \,|\, {\rm c}^-, {\rm b}^+ ; \omega_0, \omega_1)^{N({\rm e}^+, {\rm c}^-)}  \nonumber \\
    = & (\omega_0 + \omega_1 - \omega_0\omega_1)^{N({\rm e}^+, {\rm c}^+)} \omega_0^{N({\rm e}^+, {\rm c}^-)}.
\end{align}

While Eq.~\ref{appeq:likelihood-graph1-original} is not analytically tractable, it can be effectively approximated using Monte Carlo simulations. With uniform priors on~$\omega_0$ and~$\omega_1$, a reliable estimation of~$P(\mathcal{H} \,|\, G_{i \rightarrow k})$ can be obtained by generating~$m$ samples of~$\omega_0$ and~$\omega_1$ from a uniform distribution spanning the interval $[0,1]$, followed by computation of:
\begin{align}
     P(\mathcal{H} \,|\, G_{i \rightarrow k})  \nonumber = & \frac{1}{m} \sum_{i=1}^m P_1(\mathcal{H} \,|\, \omega_{0i}, \omega_{1i},  G_{i \rightarrow k}) \nonumber \\
    = & \frac{1}{m} \sum_{i=1}^m \prod_{e, c} P_1(e \,|\, c, b^+ ; \omega_{0i}, \omega_{1i})^{N(e, c)} \nonumber \\
    = & \frac{1}{m} \sum_{i=1}^m (\omega_{0i} + \omega_{1i} - \omega_{0i}\omega_{1i})^{N({\rm e}^+, {\rm c}^+)} \omega_{0i}^{N({\rm e}^+, {\rm c}^-)}  \nonumber \\
    = & \frac{1}{m} \sum_{i=1}^m (\omega_{0i} + \omega_{1i} - \omega_{0i}\omega_{1i})^{N(y_{n+1}^k=1, y_n^i=1)} \ \omega_{0i}^{N(y_{n+1}^k=1, y_n^i=0)}.
\end{align}

%% file: tables/app-graph-comparison-results.tex
\begin{table}[t]
\centering
\caption{Comparison between ground-truth graphs and inferred graphs in Junyi15 dataset. \textbf{pre} indicates evaluation against a prerequisite graph and \textbf{sim} evaluation against a similarity graph. }
\label{apptab:graph-comparison-results}
\scriptsize
    \begin{tabular}{llccccccc}
    \toprule
     &  & \acro{DKT} & \acro{DKTF} & \acro{HKT} & \acro{AKT} &  \acro{GKT} & \acro{QIKT} & \acro{PSI-KT} \\ \midrule
        \multirow{1}{*}{\begin{tabular}[c]{@{}l@{}}MRR $\uparrow$ \end{tabular}} 
            & expert pre & .0069 & .0067 & .0074 & .0075 & \underline{.0082} & .0073 & \textbf{.0086} \\ \midrule
        \multirow{3}{*}{\begin{tabular}[c]{@{}l@{}}JS $\uparrow$ \end{tabular}} 
            & expert pre & 1.46e-3 & 1.37e-3 & \underline{1.47e-3} & 1.44e-3 & 1.46e-3 & 1.19e-3 & \textbf{1.86e-3} \\ 
            & crowd pre & 4.60e-3 & 4.28e-3 & \underline{4.66e-3} & 4.48e-3 & 3.44e-3 & 5.21e-4 & \textbf{9.48e-3} \\
            & crowd sim & \underline{5.90e-4} & 0.00 & 0.00 & 5.18e-4 & 3.43e-3 & 0.00 & \textbf{4.66e-3} \\ \midrule
        \multirow{2}{*}{\begin{tabular}[c]{@{}l@{}}nLL $\downarrow$ \end{tabular}} 
            & crowd pre & 5.735 & 5.580 & 6.092 & 5.677 & \textbf{3.033} & 4.228 & \underline{4.106} \\ 
            & crowd sim & 6.598 & \underline{4.039} & 4.042 & 9.100 & 9.028  & 10.622 & \textbf{2.352} \\ 
     \bottomrule
    \end{tabular}
\end{table}

%% file: appendix/7_7_ablation_study.tex
\subsection{Ablation study} 
To thoroughly examine the various elements of \acro{PSI-KT}, including cognitive traits and the prerequisite graph, we executed three distinct ablation studies:
\begin{itemize}
    \item Without the graph inference (w/o graph): We omit the graph inference process and the influence of prerequisite KCs on the long-term mean. Essentially, this approach treats each KC independently.
    \item Without individual cognitive traits (w/o individual): We alter the variational inference network in this scenario to produce a uniform distribution across all learners. This change effectively removes the consideration of individual differences in learners' cognitive traits.
    \item Without dynamic cognitive traits (w/o dynamics): We remove the dynamic transition distribution over the traits in the generative model. This assumes that each learner has static traits over time. 
\end{itemize}
\input{tables/app-ablation-study}

\begin{figure}[t] 
    \centering
    {\includegraphics[height=1.5in]{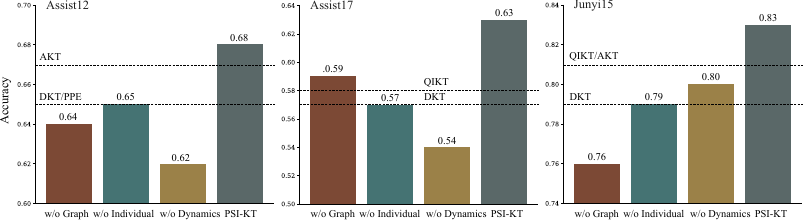}}
    \caption{Mean accuracy of \acro{PSI-KT} vs. ablations of a) the prerequisite structure (w/o graph), b) individualized learner traits (w/o individual) and c) time-dependent learner traits (w/o dynamics). Dashed lines indicate the accuracy of the two best-performing baselines.}
    \label{appfig:ablation-study}
\end{figure}

In Table~\ref{apptab:ablation-study} and Figure~\ref{appfig:ablation-study}, we present the results of our three ablation studies. We observed that the contribution of prerequisite graphs, individualized traits, and dynamic traits varied across the datasets. These findings underscore the diversity inherent in educational datasets and simultaneously reinforce the effectiveness of our unified framework.

%% file: tables/app-ablation-study.tex
\begin{table}[]
\centering
\scriptsize
\begin{tabular}{llll}
\toprule
                      & Assist12 & Assist17 & junyi15 \\ \midrule
\acro{PSI-KT}                & .68$_{.017}$  & .63$_{.015}$  & .83$_{.015}$ \\
w/o Graph             & -.04$_{.005}$  & -.04$_{.002}$  & -.07$_{.002}$ \\
w/o Individual traits & -.03$_{.002}$  & -.06$_{.006}$  & -.04$_{.001}$ \\
w/o Dynamic traits    & -.06$_{.001}$  & -.09$_{.003}$  & -.03$_{.002}$ \\ \bottomrule
\end{tabular}
\caption{The accuracy in three kinds of ablation study (mean $\pm$ \acro{SEM} across random seeds). We show the accuracy gap compared with the complete \acro{PSI-KT} model. }
\label{apptab:ablation-study}
\end{table}